\documentclass{article}

\usepackage{arxiv}

\usepackage[utf8]{inputenc} 
\usepackage[T1]{fontenc}    
\usepackage{lmodern}        
\usepackage{hyperref}       
\usepackage{url}            
\usepackage{booktabs}       
\usepackage{amsfonts}       
\usepackage{nicefrac}       
\usepackage{microtype}      
\usepackage{lipsum}
\usepackage{graphicx}

\title{Modeling the dynamics of language change: logistic regression,
Piotrowski's law, and a handful of examples in Polish}
\def\shorttitle{Modeling the dynamics of language change}

\author{
    Rafał L. Górski
   \\
    Institute of Polish Language \\
    Polish Academy of Sciences \\
  Kraków, Poland \\
  \texttt{rafal.gorski@ijp.pan.pl} \\
   \And
    Maciej Eder
   \\
    Institute of Polish Language \\
    Polish Academy of Sciences \\
  Kraków, Poland \\
  \texttt{maciej.eder@ijp.pan.pl} \\
  }

\newlength{\csllabelwidth}
\newlength{\cslhangindent}
  {}%
  {\par}
\newenvironment{CSLReferences}[2] 
 {
  \setlength{\parindent}{0pt}
  \ifodd #1 \everypar{\setlength{\hangindent}{\cslhangindent}}\ignorespaces\fi
  \ifnum #2 > 0
  \setlength{\parskip}{#2\baselineskip}
  \fi
 }%
 {}
\usepackage{calc} 

\usepackage{longtable}

\begin{document}
\maketitle

\def\tightlist{}

\begin{abstract}
The study discusses modeling diachronic processes by logistic
regression. The phenomenon of nonlinear changes in language was first
observed by Raimund Piotrowski (hence labelled as Piotrowski's law),
even if actual linguistic evidence often speaks against using the notion
of a ``law'' in this context. In our study, we apply logistic regression
models to 9 changes which occurred between 15\textsuperscript{th} and
18\textsuperscript{th} century in the Polish language. The attested
course of the majority of these changes closely follow the expected
values, which proves that the language change might indeed resemble a
nonlinear phase change scenario. We also extend the original
Piotrowski's approach by proposing polynomial logistic regression for
these cases which can hardly be described by its standard version. Also,
we propose to consider individual language change cases jointly, in
order to inspect their possible collinearity or, more likely, their
different dynamics in the function of time. Last but not least, we
evaluate our results by testing the influence of the subcorpus size on
the model's goodness-of-fit.
\end{abstract}

\hypertarget{introduction}{%
\section{Introduction}\label{introduction}}

One can hardly imagine a different picture of a language change than
that it starts among a small (possibly sociologically or spatially
restricted) group of speakers, gradually gaining more and more users
over time, even if conservative speakers somewhat retard the change.
After the first period of a slow dissemination, however, the change
accelerates until the newly introduced form is broadly accepted by
language users. The recessive form gradually becomes rare, and finally
dies out, partly because its adherents die as well. Certainly, it takes
time for the change to be completed.

The above scenario describes many other natural phenomena. Similar
processes can be found in biology, demography or epidemics. To give an
example: assume a cup of milk, into which a drop of yogurt
(i.e.~\emph{Lactobacillus delbrueckii ssp. bulgaricus}) is added. At the
beginning, the population of bacteria is small, yet increasing. Then the
growth of the population accelerates rapidly, due to binary fission of
the bacteria. However, at a certain point there is no more room (or
food, to be precise) for new bacteria to fission, therefore the increase
slows down and the population stabilizes. The same scenario applies to
the dynamics of an epidemic. In such a case, the analogy with language
change is even more striking, since people tend to ``infect'' other
members of the linguistic community with innovative forms. The more
users adopt the innovation, the higher the probability of the
``disease'' to further disseminate, i.e.~a speaker will reproduce an
innovative form which she/he heard recently or, even more typically, a
child will reproduce a form it encounters. Needless to say, significant
differences between biology and language still exist: while a virus
infection might be caused by one single contact, it is fairly naïve to
assume that language change might be caused by one attestation. Rather,
a longer exposure to the innovative form is needed for a change to
happen. We can assume with a high level of confidence that when a newly
introduced form becomes predominant, the chance of adopting it by a
conservative speaker increases as well.

On theoretical grounds, a language change can follow one of two main
scenarios. Firstly, it can be an innovation that affects a language
system without displacing any already-existing forms. In general,
vocabulary growth falls into this category (e.g.~the emergence of the
word \emph{nettiquette} in the system does not force the word
\emph{etiquette} to withdraw). The second scenario takes place when a
new form inevitably replaces an old one, e.g.~the degree to which the
form \emph{burned} (the past form of the verb \emph{to burn}) spreads in
English, by definition equals to the withdrawal of the irregular form
\emph{burnt}. In mathematical terms, the probability of finding a new
form and an old form (denoted as \emph{n} and \emph{o}, respectively)
in such a scenario is \(p(n) = 1 - p(o)\), which also implies that
\(p(o) = 1 - p(n)\). Consequently, the joint frequency of the two forms
might remain constant over the centuries, while their mutual proportions
usually vary to a significant degree.

The present study is aimed at measuring the above proportions, in order
to model the transition from the recessive to the innovative form. A
general framework for modeling such a transition will be logistic
regression, which in the context of historical linguistics is often
referred to as Piotrowski's law (Piotrowskaja and Piotrowski, 1974;
Best, 1983; Leopold, 2005).

The study will examine to which extent a theoretical model can explain
actual language changes. The examples will be drawn from the Polish
language. Namely, a few changes that are reported in handbooks of the
history of Polish to have taken place between the 15\textsuperscript{th}
and 18\textsuperscript{th} centuries, will be used to validate the
theoretical models. Even if the changes in question have been already
observed by diachronic linguistics, the present paper will rely on
substantially broader empirical material drawn from a diachronic corpus
of Polish of \emph{ca}. 12 million words, covering the time span of a
few centuries. Unlike anecdotal evidence based on single attestations
known to the linguists, a corpus-driven approach allows for harvesting
the material on an unprecedented scale, which, in turn, leads to fairly
precise comparisons of the goodness-of-fit between the observed data
points and the expected values predicted by a model.

More importantly, in our approach we will compare the dynamics of
different changes by analyzing them side by side, which to our best
knowledge has not been proposed before. In the previous studies, the
changes were analyzed as isolated phenomena, with a tacit assumption
that their trajectories must have been similar. We take into account
that an alternative hypothesis is also feasible, namely that the
dynamics of particular language changes might differ. Specifically, we
believe that there exist innovative forms that introduce the change
within decades, whereas some other need centuries to replace their
recessive counterparts.

\hypertarget{the-changes}{%
\section{The changes}\label{the-changes}}

The written record of Polish starts relatively late compared to most
languages of Western Europe. Leaving aside single attestations of Polish
words in Latin documents, the earliest written texts in Polish date from
the end of the 14\textsuperscript{th} century. The number of texts grows
significantly throughout the 15\textsuperscript{th} century,
nevertheless it can still be considered scarce. The number of extant
texts continues to be low also in the first half of the
16\textsuperscript{th} century, and after this point the volume of texts
rapidly accelerates. The increase will soon follow an exponential
pattern.

It is a claim broadly accepted that Polish saw some major changes at the
beginning of the 16\textsuperscript{th} century (Klemensiewicz, 1965).
Not only were a few grammatical categories swept out of the language
system (the aorist tense being an example, as well as the dual number),
but also the vocabulary was considerably expanded, let alone
stabilization of orthography. Also, roughly at the same time a major
literary figure and a prolific author Mikołaj Rej (1505--1569) wrote
several works that were to commence a tradition of Polish high-brow
literature, which always played an important role in forming the
standard variant of the Polish language. Next moment in history of a
similar impact on Polish is usually associated with the last phase of
the Enlightenment at the turn of the 18\textsuperscript{th} century. No
wonder, then, that the time span between the half of the
16\textsuperscript{th} and the third quarter of the
18\textsuperscript{th} century is commonly considered a distinct epoch
of Polish, usually labelled as the Middle Polish period (Klemensiewicz,
1965).

In the present study, we focus on a few changes that took place in
Middle Polish. These are as follows:

\begin{itemize}
\tightlist
\item
  \emph{abo} \textgreater{} \emph{albo} (`or')
\item
  \emph{wszytek} \textgreater{} \emph{wszystek} (`all', `entire')
\item
  \emph{inszy} \textgreater{} \emph{inny} (`another', `other')
\item
  \emph{barzo} \textgreater{} \emph{bardzo} (`very')
\item
  \emph{na-} \textgreater{} \emph{naj-} (a prefix, marker of the
  superlative)
\item
  \emph{-bych} \textgreater{} \emph{-bym} (marker of subjunctive
  1\textsuperscript{st} singular)
\item
  \emph{-bychmy} \textgreater{} \emph{-byśmy} (marker of subjunctive
  1\textsuperscript{st} plural)
\item
  \emph{-ir-} \textgreater{} \emph{-er-}\footnote{To be precise,
    phonologically this is a change of the phoneme /i/ \textgreater{}
    /e/ following a palatal consonant. Since palatalization is
    orthographically marked by the letter \emph{i} following a
    consonant, the process is reflected by substitution of a string of
    letters \emph{ir} with \emph{ier}.}
\item
  \emph{więtszy} \textgreater{} \emph{większy} (`larger', `bigger',
  comparative)
\end{itemize}

Arguably, the above changes are of very diverse nature. While
\emph{-ir-} \textgreater{} \emph{-er-} is a purely phonetic and regular
change, the first five items on the list are isolated changes. In the
transition of \emph{barzo} \textgreater{} \emph{bardzo} the fricative
/z/ is changed to the affricate /dz/. Both \emph{wszytek} \textgreater{}
\emph{wszystek} and \emph{inszy} \textgreater{} \emph{inny} are isolated
changes occurring in a single word, whereas \emph{na-} \textgreater{}
\emph{naj-} is a modification of a morpheme that was not caused by a
regular sound change. The transition \emph{-bych} \textgreater{}
\emph{-bym}, as well as \emph{-bychmy} \textgreater{} \emph{-byśmy}, is
a morphological change with one morpheme replacing another ({[}x{]}
\textgreater{} {[}m{]} and {[}xmi{]} \textgreater{} {[}smi{]}).

\hypertarget{the-so-called-piotrowskis-law}{%
\section{The so-called Piotrowski's
law}\label{the-so-called-piotrowskis-law}}

As already mentioned, the replacement of the recessive form by its
innovative counterpart is never immediate. Just the contrary, the two
forms coexist for some time. In probabilistic terms, it can be said that
the probability of finding the innovative form in the period before the
change is roughly 0 (tending to 0). In turn, when the change is
completed, the innovative form occurs with a probability of 1 (or
tending to 1). In the transition period, however, the proportion of the
two forms as a function of time follow an ``s''-shaped curve (see
e.g.~Figs. 1--6), which can be described by a non-linear mathematical
model. This process has been originally observed by Rajmund Piotrowski
(1974), and elaborated by Gabriel Altmann (1983; Köhler and Tuzzi,
2015):

\[p(t) = \frac{1}{1 + a \times e^{-rt}}\]

where \emph{a} is an arbitrary parameter -- a positive number by which
the trajectory of the function is shifted on the time axis; \emph{t} = 0
when the proportion between recessive and innovative forms is 1:1; and
\emph{r} is a parameter which is to be adjusted to the empirical data.
Traditionally, \emph{e} denotes the Euler's number. The parameter
\emph{r} allows for shaping the curve, that is to stretch it to a
desired extent.

Interestingly, the same model can describe two different linguistic
phenomena. Not only is it applicable to language changes, but also to
orthographic variation (Basu, 2016) and most notably, to lexical
borrowings: loanwords from a certain donor percolate to the host
language in a timespan of the donor's cultural attractiveness. The
number of loanwords originating from one language-borrower in the host
language is initially small, and then it raises gradually until it
reaches a point of stabilization, with virtually no new loanwords, when
the donor language is no more fashionable or the contacts are less
intensive than before (Best, 2013; Stachowski, 2016; Gnatchuk, 2015).

For anyone familiar with statistic methodology it is rather obvious that
the ``s''-shaped curve is usually associated with logistic regression
models. And indeed, it has been observed (Vulanović, 2007; Vulanović and
Baayen, 2007; Köhler and Tuzzi, 2015) that the model suggested by Altman
and Piotrowski is, in fact, a variant of the well-known logistic
regression model:

\[ p(X) = \frac{1}{1 + e^{-(\beta_0 + \beta_1 X)}} \]

where \(\beta_{0}\) and \(\beta_{1}\) determine the logistic intercept
and slope. The major advantage of using the logistic regression over the
original Piotrowski's model is that the former allows for a direct
comparison of two variables -- the proportion of innovative form against
the year -- whereas in the original approach, the variable \emph{t} (for
time) remains relative and uses a somewhat vague numeric scale,
e.g.~from -6 to 6. Another reason that makes logistic regression a
better choice over the original Piotrowski's equation is its simpler
form which requires less parameters. {In practical terms, logistic
regression is a time-proven solution, as it is routinely used in
countless applications in quantitative linguistic and beyond, and it can
be performed using out-of-the-shelf statistical software.}

{Last but not least, logistic regression models are able to account for
data imbalance. Given the almost-exponential growth of literary
production over the centuries, one can safely assume that the number of
attestations retrieved from medieval works will be significantly
scarcer than the evidence from the later periods. While logistic
regression is fairly resistant to data imbalance, the extreme unevenness
of diachronic corpora might prove problematic. The issue might or might
not be important (it has never been raised in previous studies, let
alone systematic testing), therefore we will report the results for both
standard logistic regression models, and their weighted variants. As
prior weights, we use the sum of innovative and recessive forms'
attestations in each time span.}

\hypertarget{corpus-and-text-preprocessing}{%
\section{Corpus and text
preprocessing}\label{corpus-and-text-preprocessing}}

The study is based on a tailored diachronic corpus of Polish, which
covers a period between 1380 and 1850 and comprises \emph{ca}. 12
million running words in 790 texts (Górski et al., 2019).\footnote{Due
  to copyright restrictions, the corpus could not be made publicly
  available. However, we post all the derivative datasets used in this
  study, as well as the full set of the results, followed by the code
  needed to replicate the tests, on our GitHub repository:
  \url{https://github.com/computationalstylistics/PL_language_change}}
While the size of the corpus was essential for the purpose of the
current study, our corpus was compiled with an eye on its bulk rather
than balance. Understandably, in the case of early historical data the
number of available texts is by definition limited, moreover the entire
textual output of the Polish Middle Ages consists of religious and legal
texts. A simple rule applies here: the older the time period to be
covered by a corpus, the bigger the expected bias. Therefore, we were
guided by the principle that any text extant was included in the corpus,
as long as it was available in electronic format.

Similarly, the distribution of the texts over time was uneven: in terms
of running words, some 5\% of the corpus (or 0.72 million running words)
covers the time until 1550. In turn, the years 1550--1600 make up 12\%,
the 17\textsuperscript{th} century -- 47\%, the 18\textsuperscript{th}
century -- 27\%, and finally 9\% of the corpus covers the years
1800--1850. We decided not to diminish the disproportionately large
subcorpus of the 17\textsuperscript{th} century. However: since our
study is aimed at examining the \emph{proportion} between two forms -- a
recessive one and an innovative one -- and not at actual word
occurrences over time, uneven temporal coverage of the corpus should not
mislead a logistic regression model to a significant extent. Also, it
can be seen that the two outermost periods (before 1550 and after 1800)
are significantly smaller than the others. These periods, however, in
most cases serve as a background, or the time when either the change has
not started yet, or it has been already completed. Still, one has to
bear in mind that the corpus includes the entire known textual output of
the Middle Ages, and nearly so for the first half of the
16\textsuperscript{th} century.

A potentially severe issue is an uneven representation of certain text
genres in particular time periods, e.g.~the overrepresentation of
religious treatises in the late Middle Ages, as well as the
overrepresentation of \emph{belles lettres} in the
19\textsuperscript{th} century. Regardless of several reasons of such a
textual/cultural bias in any diachronic corpus, we hardly believe it can
ever be reliably corrected for. However, since different genres do not
affect morphology as they affect lexis, we assume that grammatical
language changes will be reliably reflected in our corpus anyway. We did
not make use of POS tagging of the texts, and again for the same reason.
A bulk of research shows that the accuracy of tagging for historical
texts is substantially lower compared to modern ones; the same rule is
probably even more valid in the case of under-resourced languages, such
as Polish. Luckily enough, however, the language phenomena to be
discussed in the present study can be easily accessed via plain text
queries, in some cases (e.g., \emph{-ir-} \textgreater{} \emph{-ier-})
followed by manual final selection.

Yet another caveat needs to be introduced. The texts collected in the
corpus might be, and usually are, of different size. We did not attempt
at trimming long texts, though. Firstly, it is widely agreed upon in
corpus linguistics that the texts should be included as a whole rather
than sampled, since each part of a text has its own peculiarities
(Meyer, 2002). Secondly, with the scarcity of historical data, it would
be improvident to let large amounts of already-acquired data be wasted.
At the same time, the cost we have to accept when taking entire texts
rather than sampling cannot be neglected. Namely, if a long text is more
conservative (or more progressive) than its contemporaries, such a
single text can skew the results by shifting the curve away from the
general tendency. The size of the effect might differ depending on the
frequency of the overrepresented form in question.

Finally, it should be kept in mind that the final shape of the texts is
a product of philologists. The linguist can only trust their
meticulousness and expertise, by accepting their work with all its
merits and potential flaws.

In order to let the reader get an idea of the number of attestations for
the changes under scrutiny:

\begin{itemize}
\tightlist
\item
  \emph{abo} \textgreater{} \emph{albo}: 44,743
\item
  \emph{wszytek} \textgreater{} \emph{wszystek}: 55,005
\item
  \emph{inszy} \textgreater{} \emph{inny}: 25,103
\item
  \emph{barzo} \textgreater{} \emph{bardzo}: 17,938
\item
  \emph{na-} \textgreater{} \emph{naj-}: 8,832
\item
  \emph{-bych} \textgreater{} \emph{-bym}: 7,491
\item
  \emph{-bychmy} \textgreater{} \emph{-byśmy}: 2,585
\item
  \emph{-ir-} \textgreater{} \emph{-er-}: 49,829
\item
  \emph{więtszy} \textgreater{} \emph{większy}: 9,158
\end{itemize}

{Certainly, the above attestations are unevenly distributed over the
centuries; they are generally scarce in the 15th century and denser in
subsequent periods.}

\hypertarget{method}{%
\section{Method}\label{method}}

Modeling a language change by logistic regression requires the input
data to be split into ``bins'', or subcorpora spanning particular time
windows in which the proportion of the recessive and the innovative form
is being measured. In fact, this means that the input corpus has to be
divided into a series of sequentially ordered subcorpora. Depending on
the corpus, a natural time window size could range from one day (in the
case of contemporary monitor corpora) to, say, one century. Since the
texts collected in our corpus are marked for the publication year, it
might seem convenient to decide on one-year chunks. We used wider bins,
however, and for a few reasons. Firstly, the aim of our study was to
observe changes in language, rather than changes between particular
texts; aggregating lumps of texts into bins would account for this
factor. Secondly, the alleged one-year precision in our metadata happens
to be misleading, especially when
15\textsuperscript{th}/16\textsuperscript{th}-century texts are
concerned. In some cases the dating relies on very rough estimation,
ranging from a few years to a few decades, while some other texts are
known to be, say, 16\textsuperscript{th}-century copies while the
original text must have been written earlier. Also, there are some texts
that were created over a long period of time. Dividing the original
corpus into bins partially mitigates the metadata imperfections.

Regardless of the chosen size of the bins, some side effects cannot be
ruled out. Assume there exists a corpus that has been split into 20-year
subcorpora. Now assume there exist three texts written in 1602, 1618 and
1622, respectively. The division rule makes the first two texts fall
into one bin (1601--1620), and the third one to another (1621--1640),
even if the first two texts are separated from each other by 16 years,
whereas the second and the third one by 4. Note that the larger the time
spans covered by the subcorpora, the bigger the unwanted effect. In
order to avoid the above issues, we involved a ``moving window''
procedure, in which the subsequent bins were excerpted with an overlap.
Not only does it allow for more data points, but it also diminishes the
effect of Procrustes's bed of introducing arbitrary borderlines between
subcorpora. In the aforementioned example, the text from 1618 would
still fall into the subcorpus 1601--1620 together with the text from
1602, but additionally into the second subcorpus 1611--1630 with the
text from 1622. The advantages of the ``moving window'' procedure by far
surpasses its downsides, which include the fact that a single outlier
affects more than one bin, and thus more than one data point.

Consequently, two parameters are used in our study: (i) the size of the
time window, and (ii) the overlap of adjacent windows. Below, we present
the results obtained for 20-years' windows, with a 10-year overlap.
Since the choice of both parameters can affect the results, we
systematically tested other window sizes; the results are reported
below, in the Evaluation section. In order to cross-check the results,
we also tested the behavior of one-year chunks without any moving
window.

To conduct the tests discussed in the present paper, we used the R
programming environment with a custom function for dividing the dataset
into bins by the moving window procedure. The logistic regression models
were fitted using a standard R function to compute General Linear
Models, namely \texttt{glm()}, combined with a logit kernel and -- if
applicable -- with prior weights. All the datasets used in this study,
together with the final plots and an R script to replicate the results,
are posted on our GitHub repository:
\url{https://github.com/computationalstylistics/PL_language_change}.

\hypertarget{results}{%
\section{Results}\label{results}}

\hypertarget{wiux119tszy-wiux119kszy}{%
\subsection{\texorpdfstring{\emph{więtszy} \textgreater{}
\emph{większy}}{więtszy \textgreater{} większy}}\label{wiux119tszy-wiux119kszy}}

The first examined transition is \emph{więtszy} \textgreater{}
\emph{większy} (in their all cases, genders and numbers, including the
superlative, i.e.~\emph{na(j)więtszy}). Fig. 1 shows the proportion of
recessive to innovative forms or, to be more precise, the probability to
find an innovative form \(p(i)\) in a given chronologically ordered
subcorpus. The first occurrence of the innovative form is found in a
text dated 1543, whereas the last recessive form is attested as late as
in 1825. However, no occurrences of the recessive forms appear between
1746 and 1825.

Fig. 1 shows clearly that the mutual relations of \emph{więtszy} and
\emph{większy} (i.e.~their proportions in individual subcorpora) are
arranged as if they were following an invisible curve. There is no doubt
that we are dealing with a regularity here. This hypothetical curve --
being actually a sequence of local predictions inferred from the
empirical data -- is precisely the model we are looking for. The
relation of the curve (model) to the points (data) shows the very aim of
modeling: it is an attempt to understand complex processes by
identifying a hypothetical line that is optimally fitted through the
empirical data points. The underlying assumption is that the
regularities hidden to the human eye are responsible for ``generating''
actual phenomena (with all their imperfections) and therefore it is
possible to decode the original regularities by a model -- a retrofitted
idealization to empirical data.

\begin{figure}
\centering
\includegraphics[width=1\textwidth]{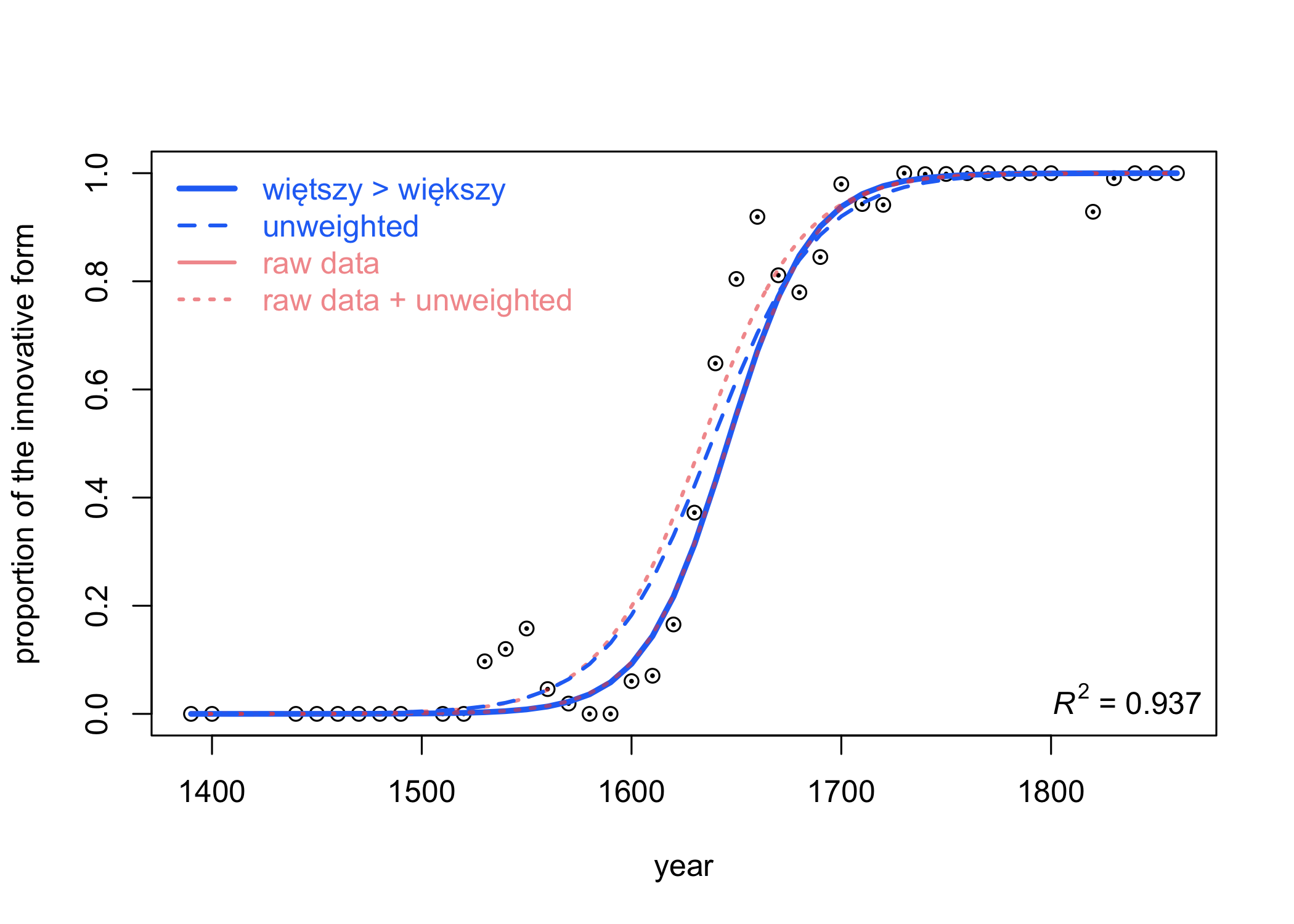}
\caption{The change \emph{więtszy} \textgreater{}
\emph{większy}. The points show the proportion between the recessive and
innovative form (empirical data), the solid line presents a fitted
logistic model with prior weights, whereas the dashed line -- an
unweighted model (i.e.~simple logistic regression). Two other lines show
models inferred from a raw dataset, discretized at the level of one
year.}
\end{figure}

{The plot shows four logistic regression models, denoted with solid,
dashed and dotted lines. They appear to be similar yet it is not obvious
at all which of them optimally describes the original dataset.} Before
we can claim any of the models valid, their statistical significance
needs to be assessed. To this end, we will be using standard diagnostic
tools: the \emph{p}-value as a measure of statistical significance, and
\emph{R}\textsuperscript{2} for assessing the goodness-of-fit. Since the
original \emph{R}\textsuperscript{2} measure is shown to be misleading
in logistic regression setups, {we use a
pseudo-\emph{R}\textsuperscript{2} variant as proposed by McFadden
(1973). Whenever we report any \emph{R}\textsuperscript{2} values
hereafter, we mean McFadden's pseudo-\emph{R}\textsuperscript{2}
measure.}

In case of the change \emph{więtszy} \textgreater{} \emph{większy}, the
data points follow the idealized curve very closely, as if the curve was
attracting them -- arguably, the logistic regression describes the
observed data very well. This intuition is corroborated by a very high
statistical significance, \(p < 0.001\), under the optimal parameters of
the model (\(\beta_0 = -82.238\), \(\beta_1 = 0.049\), at 42 degrees of
freedom). The model is presented in Fig. 1 with a solid line. The
goodness-of-fit, or the match of the logistic curve with the observed
data, is also very high, with \emph{R}\textsuperscript{2} = 0.937.

{The difference between the weighted model (solid line) and an
unweighted one (dashed line) line is very small, yet noticeable. With
the inferred parameters \(\beta_0 = -64. 569\), \(\beta_1 = 0.394\), the
unweighted model exhibits the goodness-of-fit only marginally worse
compared to the weighted version (\emph{R}\textsuperscript{2} = 0.932
vs.~0.937). The potential bias introduced by imbalanced input data does
not appear to be significant, at least in this case (some advantages of
using weighted models will be discussed later, see Fig. 4--5).}

Certainly, a question might be asked whether a different size of the
subcorpora and/or a different overlap of the moving window could have
affected the results. Similarly, one might ask whether a model inferred
from raw data -- presumably as fine-grained as possible -- would exhibit
a different picture. While the former question will be discussed in the
Evaluation section, here we briefly address the latter one. The dashed
line in Fig. 1 represents a logistic regression model fitted to the raw
dataset. Not only is the model statistically significant (\(p < 0.001\)
at 200 degrees of freedom), but it is virtually indistinguishable from our base model involving bins (while the
unweighted version follows a slightly different path, see Fig. 1, dotted
line). It is true that the \emph{R}\textsuperscript{2} = 0.669 is much
lower than in the base model, but this is due to the fitting conditions:
a big number of degrees of freedom by definition produces more residuals
and inevitably deteriorates \emph{R}\textsuperscript{2}. For this
reason, the value 0.0.669 cannot be directly compared to all the other
\emph{R}\textsuperscript{2} scores discussed in the paper. Even though,
the obtained goodness-of-fit for such a raw model can be claimed high.

\hypertarget{barzo-bardzo}{%
\subsection{\texorpdfstring{\emph{barzo} \textgreater{}
\emph{bardzo}}{barzo \textgreater{} bardzo}}\label{barzo-bardzo}}

The second change is \emph{barzo} \textgreater{} \emph{bardzo}. Unlike
other cases under our scrutiny, this is a relatively late change -- it
is virtually impossible to spot any attestations of the newer form until
the early 17\textsuperscript{th} century. In our corpus, we noted only
25 occurrences of \emph{bardzo} (not only in positive degree, but also
in comparative as well as in superlative) with over 2,213 occurrences of
the form \emph{barzo} (and \emph{barziej} and \emph{najbarziej}) before
the year 1600. From that moment on, the process of transition slowly
accelerates; the rapid increase in the share of the innovative
\emph{bardzo} occurs at the end of the 18\textsuperscript{th} century.
{As shown in Fig. 2, the process is well described by logistic
regression, with a high goodness-of-fit \emph{R}\textsuperscript{2} =
0.815, and even a higher value of 0.919 for the unweighted (i.e.~simple)
logistic regression model.}

However, a deviation from the model in the second half of the
18\textsuperscript{th} century cannot be neglected. Two outliers that
clearly differ from the modeled course of the change perhaps could be
attributed to our data collection procedure: we have computed the
frequencies collectively for the positive, comparative and superlative
degree, while it could have happened that positive (\emph{barzo}
\textgreater{} \emph{bardzo}) and comparative as well as
superlative\footnote{Note that in Polish, comparative and superlative
  share the same suffix, while the latter differs from former by the
  prefix \emph{naj-}.} changed at a different pace. Also, it is possible
that a single author -- particularly conservative in this case -- was
responsible for this effect.

\begin{figure}
\centering
\includegraphics[width=1\textwidth]{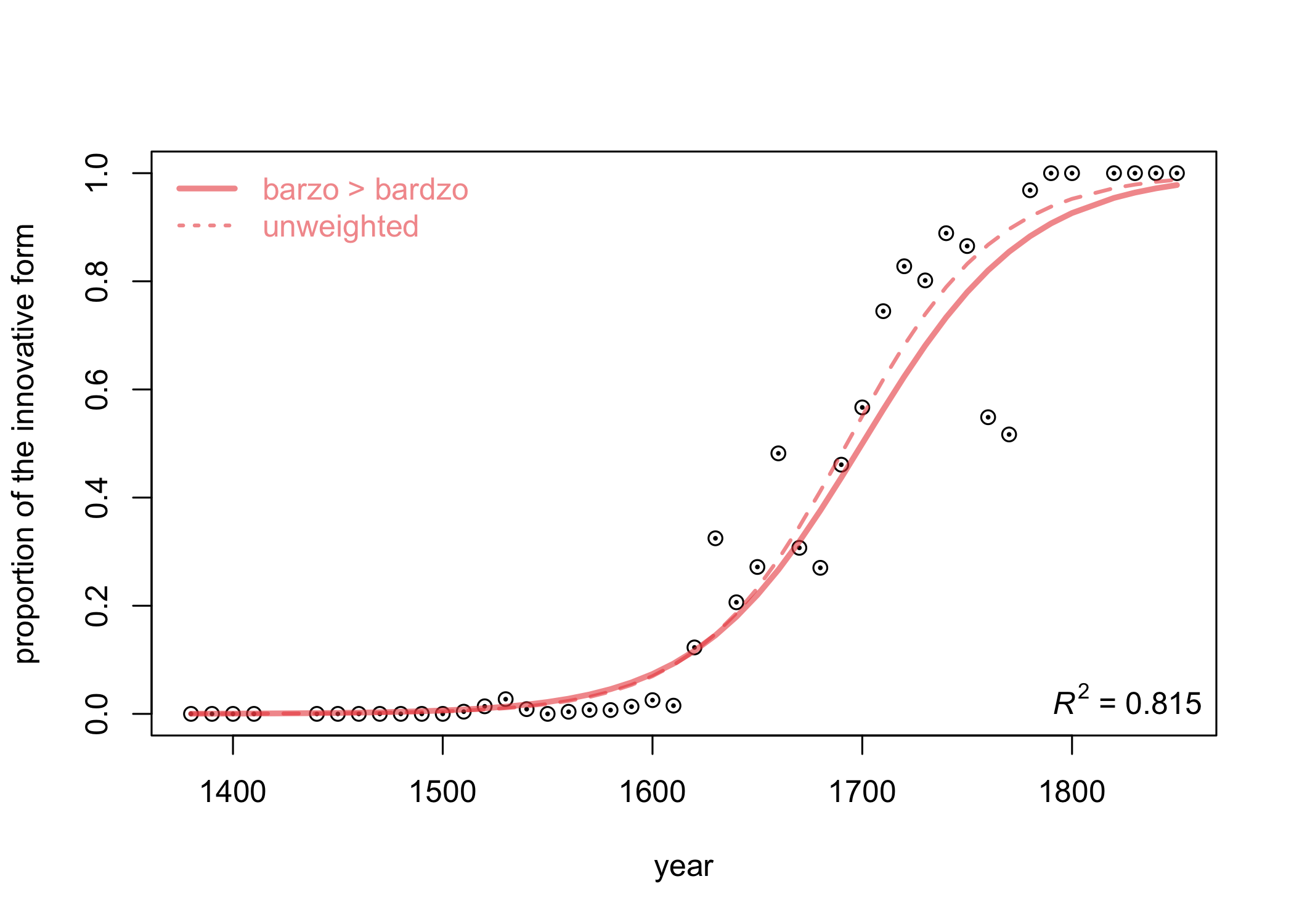}
\caption{The change \emph{barzo} \textgreater{} \emph{bardzo}.}
\end{figure}

\hypertarget{bych--bym--bychmy--byux15bmy}{%
\subsection{\texorpdfstring{-\emph{bych} \textgreater{} -\emph{bym,
-bychmy} \textgreater{}
\emph{-byśmy}}{-bych \textgreater{} -bym, -bychmy \textgreater{} -byśmy}}\label{bych--bym--bychmy--byux15bmy}}

The late 15\textsuperscript{th} and early 16\textsuperscript{th} century
witnessed the coexistence of \emph{-bym, -byśmy} (the forms used in the
present day Polish) and \emph{-bych}, \emph{-bychmy} as markers of the
1\textsuperscript{st} person singular and plural,
accordingly\footnote{To be precise, apart from the above mentioned there
  are also markers of dual: \emph{-swa} and \emph{-chwa}. However, since
  dual died out in the early 16\textsuperscript{th} century, we will
  gloss over these morphemes.}. The competition of these two forms was
studied, among others, by Taszycki (1946), Kowalska (1978), as well as
Motyl (2014). Taszycki links the forms \emph{-bych, -bychmy} with the
region of Lesser (southern) Poland, since authors from elsewhere used it
rarely. It was also noticed that the course of the change of the two
morphemes was different -- it persisted longer in the plural than in the
singular.

Fig. 3 shows the course of change of the markers of both numbers
together. The first thing that immediately catches the eye is the very
good fit of the model to the data: -\emph{bych} \textgreater{}
-\emph{bym}, -\emph{bychmy} \textgreater{} -\emph{byśmy} is a perfect
example of logistic regression. The value of \emph{R}\textsuperscript{2}
= 0.0.957 (or 0.976 for the unweighted model) at $p \approx 0$ only
confirms the observations made by eyeballing.

\begin{figure}
\centering
\includegraphics[width=1\textwidth]{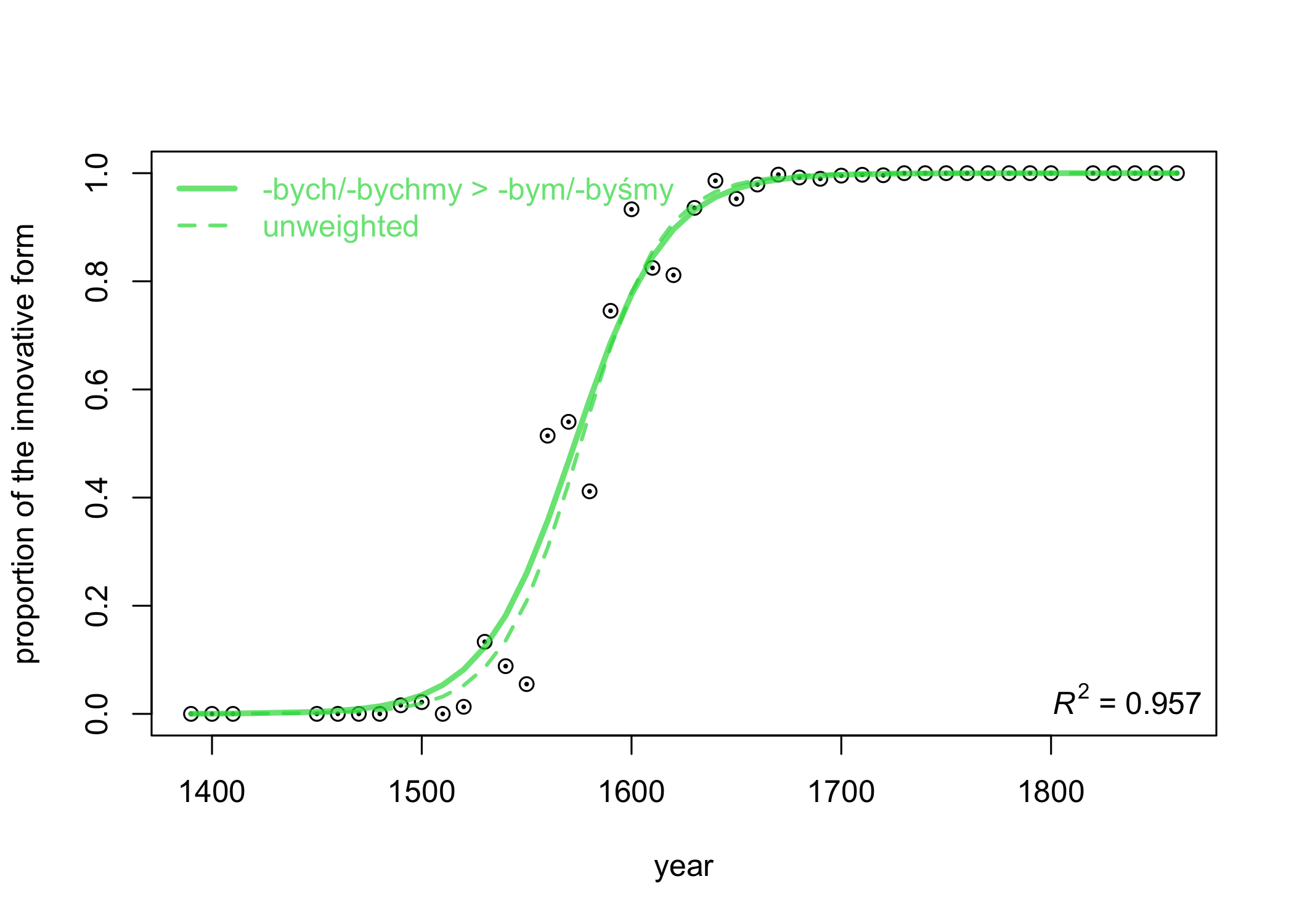}
\caption{The course of change of \emph{-bych}/\emph{-byśmy}
\textgreater{} \emph{-bym}/\emph{-byśmy}.}
\end{figure}

Two forms of the same verb -- namely 1\textsuperscript{st} person
singular and 1\textsuperscript{st} person plural -- can be modeled
together, as shown in Fig. 3, but they can be also scrutinized
individually. Fairly plausible is a scenario in which one of the forms
gains its popularity sooner, whereas the other form is following up at a
slower pace. Therefore, we also computed independent models for the two
forms (i.e.~singular and plural). While the model for \emph{bych
\textgreater{} bym} alone proved to be virtually indistinguishable from
the above combined approach, the plural form was different: the
beginning of the change turned out to be blurry and {therefore difficult
to fit a logistic regression model. However, here we face a situation
where making a model aware of data imbalance -- by adding weights to the
equation -- improves the performance significantly. The weighted model
not only exhibits a higher goodness-of-fit \emph{R}\textsuperscript{2} =
0.618 (vs.~0.558 for the unweighted variant), but it also resembles the
``s''-curve to a greater extent (Fig. 4, solid vs.~dashed line). The
weighting effect can be appreciated with a naked eye: in Fig. 4, we
log-scaled the data points according to the number of attestations in
particular bins. While in the period 1450--1460 we deal with 7
occurrences of both forms, in the time span 1600--1610 the number of
attestations reaches 451. Apparently, the weighted model pays less
attention to the faint data points (including the outliers!) than to the
opulent ones. If we assume that the textual evidence from early
documents is on average less reliable than later sources, weighted
logistic regression models might indeed be preferred over the classical
models, let alone the original Piotrowski's equation.}

\begin{figure}
\centering
\includegraphics[width=1\textwidth]{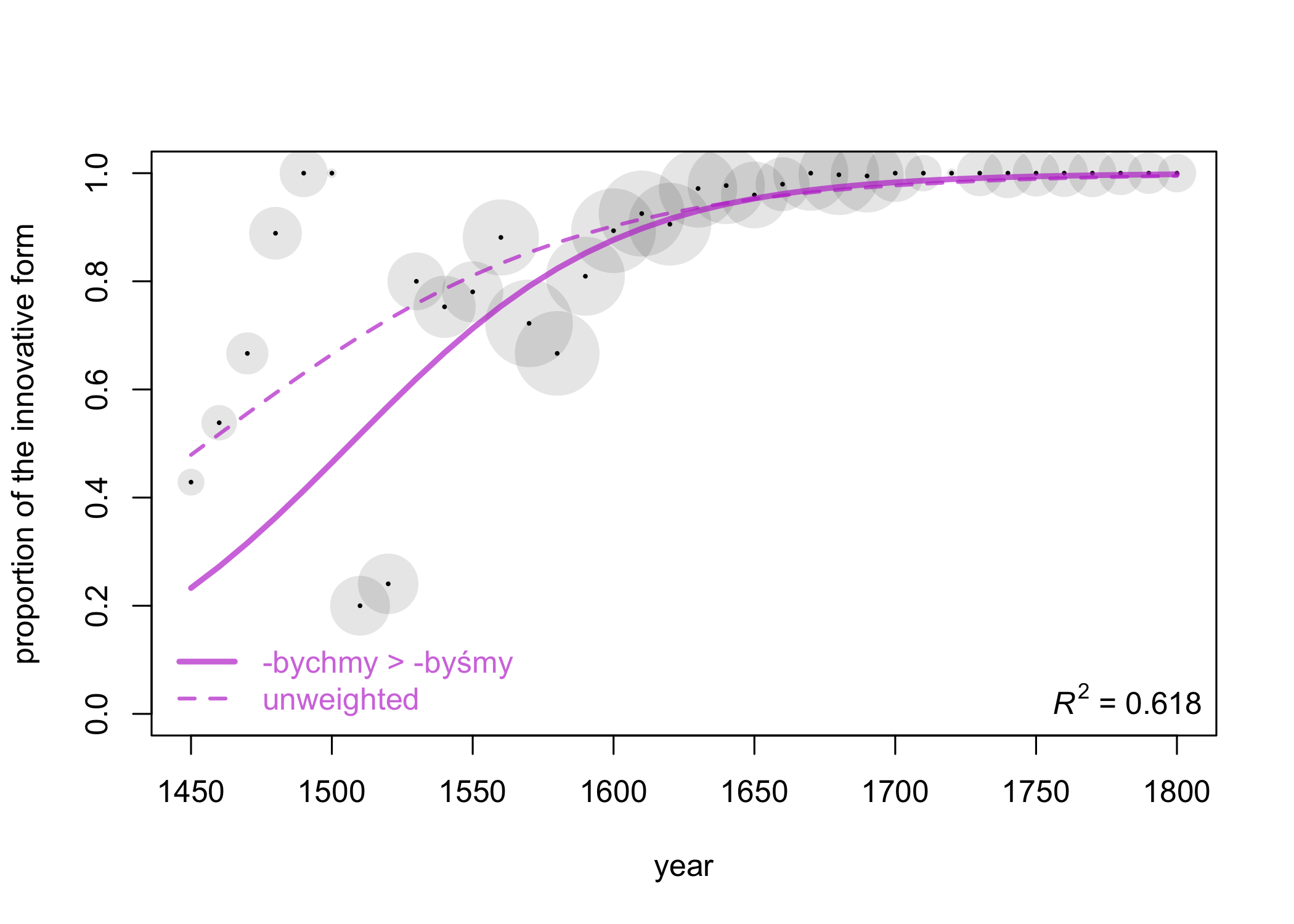}
\caption{The course of change of the form \emph{bychmy}
\textgreater{} \emph{byśmy} alone. The size of the data points is the
natural logarithm of the number of attestations in each time span.}
\end{figure}

\hypertarget{na--naj-}{%
\subsection{\texorpdfstring{\emph{na}- \textgreater{}
\emph{naj}-}{na- \textgreater{} naj-}}\label{na--naj-}}

Disregarding very few (and uncertain) occurrences of \emph{naj}- in the
15\textsuperscript{th}-century part of our corpus, the first attestation
of the innovative form can be dated 1535, while its recessive
counterpart is noticed for the last time in 1772. As far as the
15\textsuperscript{th}-century period is concerned, the data points are
highly blurred. Moreover, due to the overall scarcity of data in the
15\textsuperscript{th} century, the marginal occurrence of the
innovative form in two psalters significantly increases the proportions
of the innovative to the recessive form. The course of the change in the
entire corpus is illustrated in Fig. 5.

\begin{figure}
\centering
\includegraphics[width=1\textwidth]{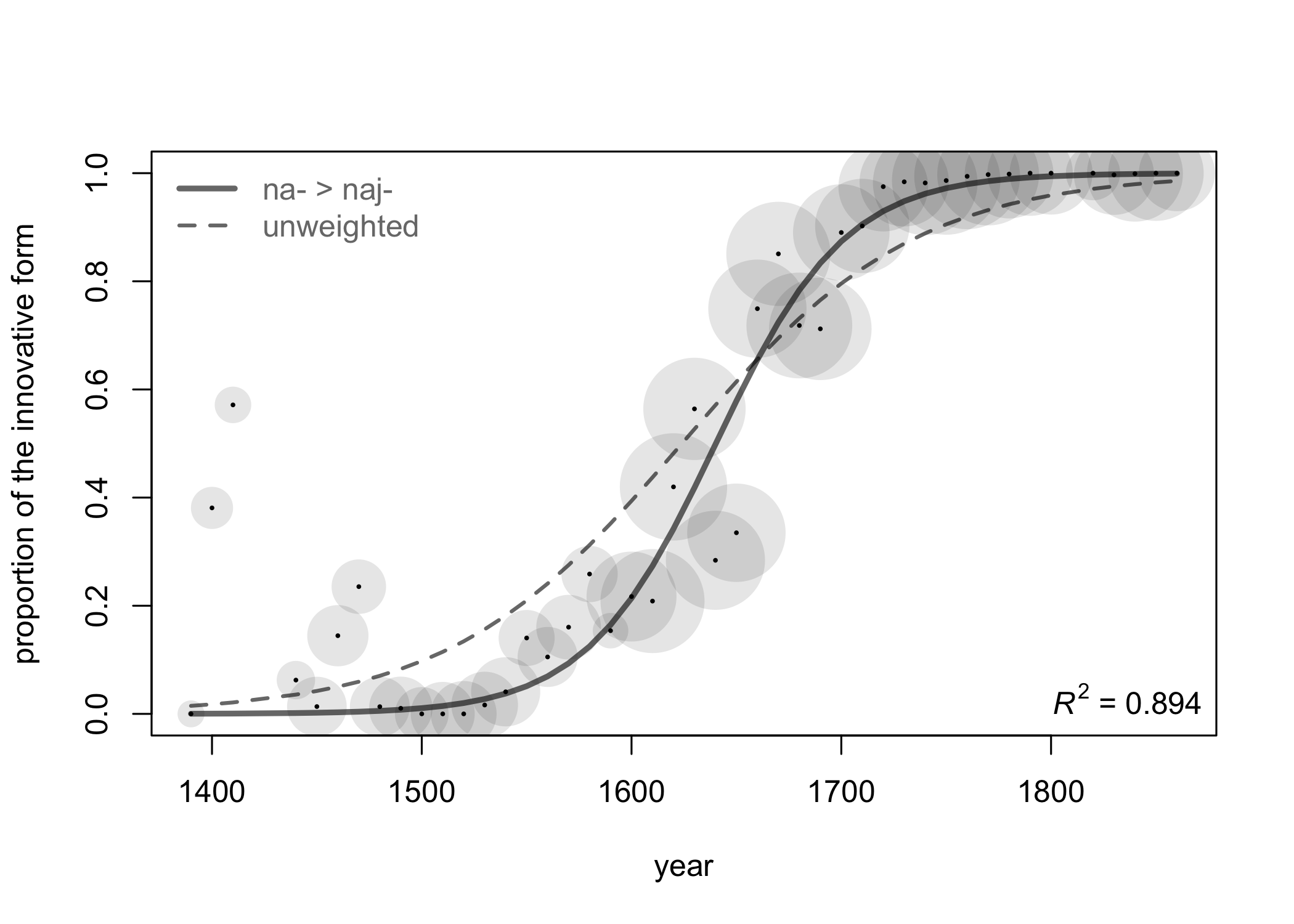}
\caption{The course of change of the superlative marker
\emph{na-} \textgreater{} \emph{naj-}. The size of the data points is
the natural logarithm of the number of attestations in each time span.}
\end{figure}

Despite these blurred data points in the oldest Polish texts, the fit of
the theoretical course of logistic regression to empirical data is not
as bad as one might think, with \emph{R}\textsuperscript{2} = 0.818 at
\emph{p} \textless{} 0.001 {(Fig. 5, dashed line). Here, the weighted
variant of our model again proves superior, with its significantly
higher \emph{R}\textsuperscript{2} = 0.894, and visually better
trajectory (solid line). It appears, once more, that an imbalanced
dataset can be corrected for with a weighted model.}

\hypertarget{ir---er-}{%
\subsection{\texorpdfstring{-\emph{ir}- \textgreater{}
-\emph{er}-}{-ir- \textgreater{} -er-}}\label{ir---er-}}

Fig. 6 depicts the competition of the forms with recessive -\emph{ir}-
and innovative -\emph{er}-, such as e.g.~\emph{śmirć} \textgreater{}
\emph{śmierć} `death', \emph{cirpieć} \textgreater{} \emph{cierpieć} `to
suffer', \emph{pirwej} \textgreater{} \emph{pierwej} `first, before'. As
can be seen in the plot, this change is an imperfect example of the
logistic process, since the corpus does not reveal the beginning of the
change -- the innovative form can be found already in the first
subcorpus. The proportion of the innovative form raises consistently.
Another untypical feature is a very long period when the recessive forms
are barely attested. It seems that the process of this change was
completed in the beginning of the 17\textsuperscript{th} century, but
the archaic forms occasionally emerged in the course of the next two
centuries. One possible explanation is that the speed of the change is
different for particular words. It is also possible that in the
18\textsuperscript{th} century there was a kind of temporary return to
the archaic form motivated by a fashion. Since the beginning of the
change remains hidden in the pre-literate era and thus invisible to the
logistic model, it comes as no surprise that the fit of the model,
\emph{R}\textsuperscript{2} = 0.619 (and 0.853 in the unweighted
variant), is lower than in the cases discussed above, yet still
acceptable.

\begin{figure}
\centering
\includegraphics[width=1\textwidth]{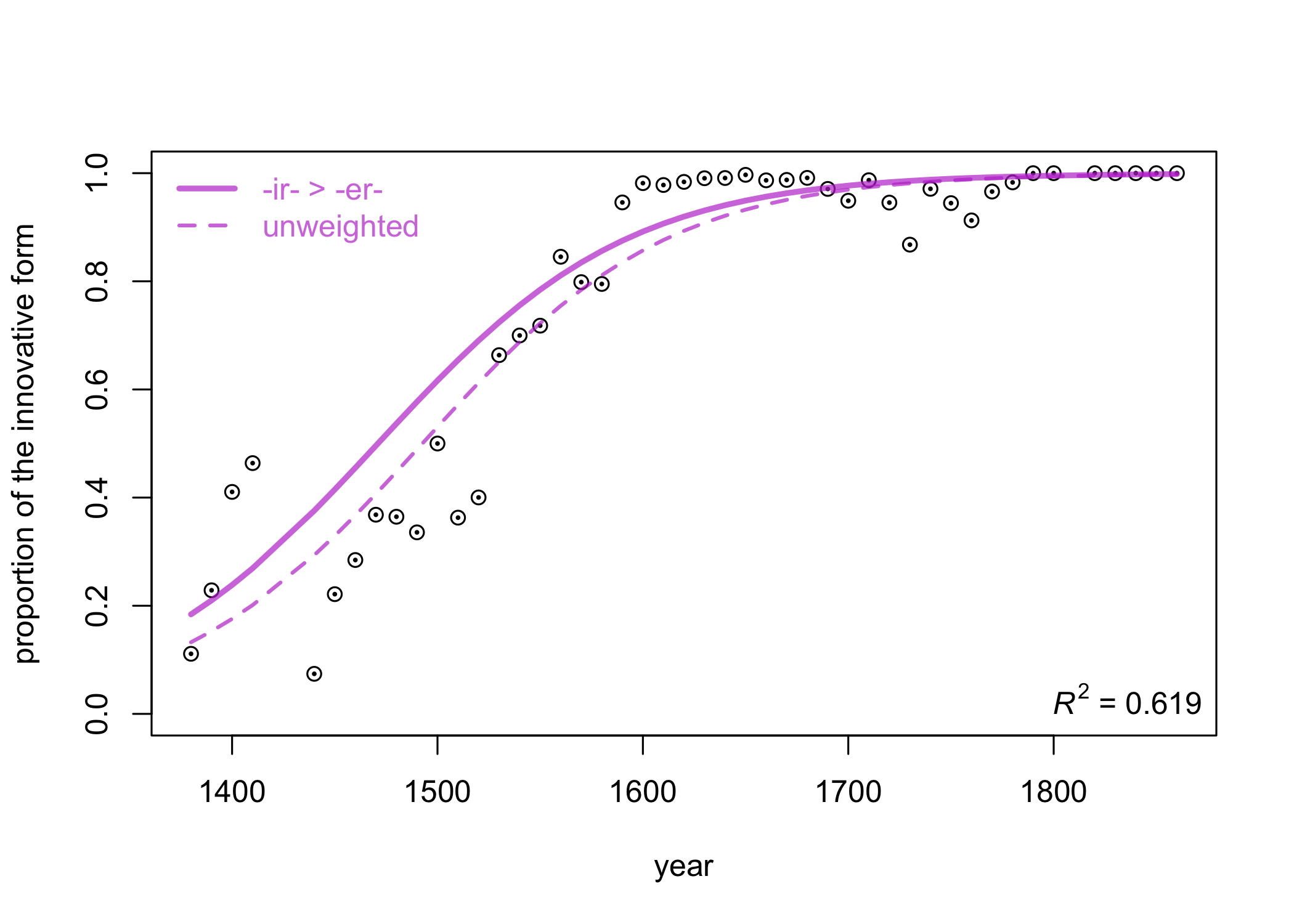}
\caption{The course of change -\emph{ir}- \textgreater{}
-\emph{er}-.}
\end{figure}

\hypertarget{inszy-inny}{%
\subsection{\texorpdfstring{\emph{inszy} \textgreater{}
\emph{inny}}{inszy \textgreater{} inny}}\label{inszy-inny}}

This change can hardly be described in terms of logistic regression in
its standard form. Should we restrict ourselves to the procedure which
we applied so far, little could be said about the nature of the process.
The goodness of fit of such a model is unsatisfactory, with
\emph{R}\textsuperscript{2} = {0.488. The performance of a weighted
model proves substantially better, with \emph{R}\textsuperscript{2} =
0.655 (Fig. 7, dotted line), yet one cannot deny the fact that the data
points hardly follow a logistic trajectory}. Two immediate conclusions
can be formulated here: firstly, Piotrowski's original idea of modeling
the change by an ``s''-shaped curve does not apply universally, and
therefore it cannot be referred to as a linguistic law. Secondly, while
standard logistic regression model proves unsatisfactory, other options
from the generalized regression family can be used instead. And indeed,
the polynomial logistic regression model (James et al., 2013: 265--300)
seems to be a good choice here. The principle of such a model, in a
nutshell, is that instead of limiting the modeled curve to the ``s''
shape, we allow for its greater flexibility. Depending on the degree of
the polynomial, the line may bend many times; each subsequent parameter
of the polynomial allows for two additional bends.

\begin{figure}
\centering
\includegraphics[width=1\textwidth]{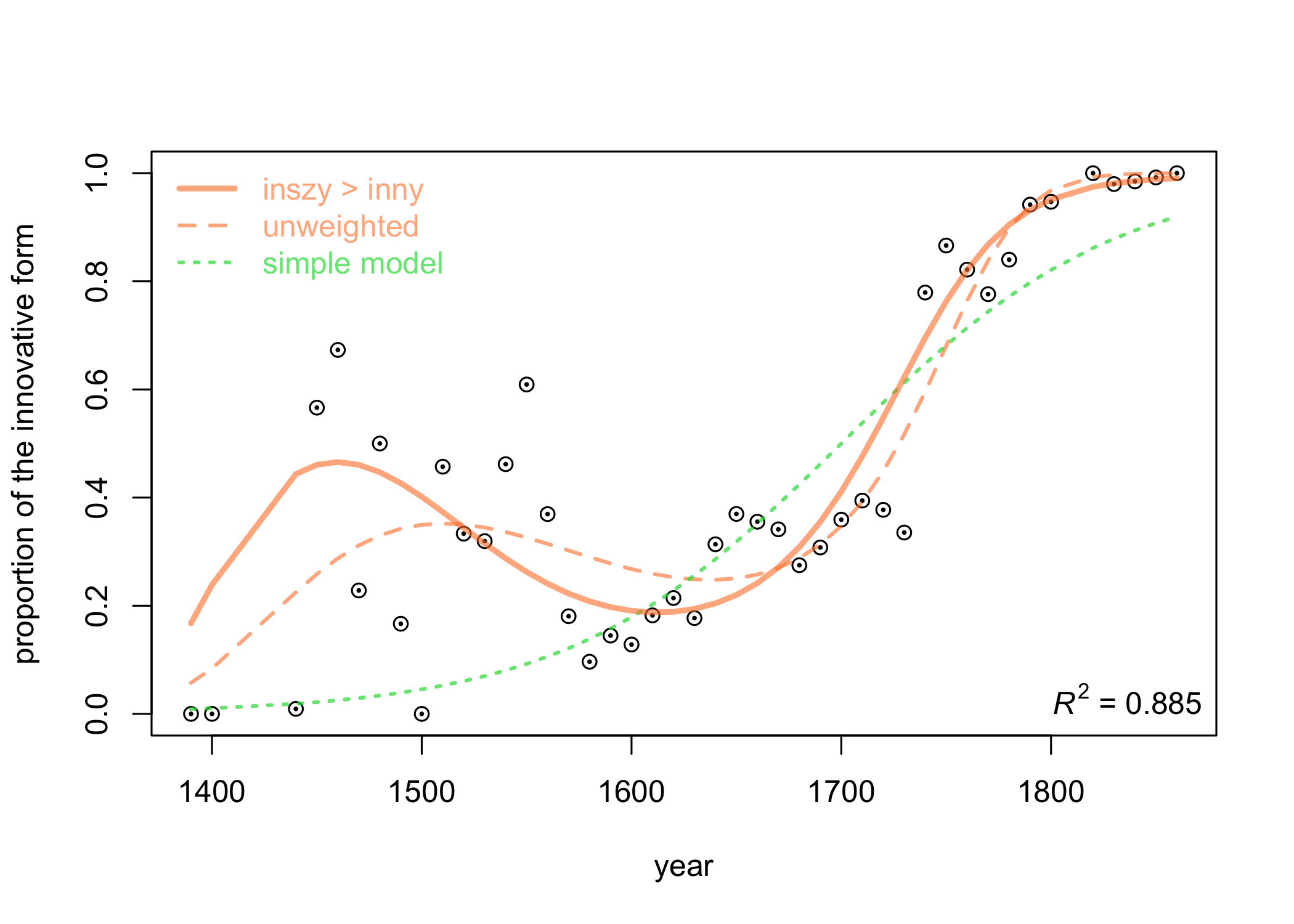}
\caption{The course of change \emph{inszy} \textgreater{}
\emph{inny} (in all inflected forms). Polynomial logistic regression
model with prior weights (solid line) and unweighted (dashed line)
vs.~simple logistic regression (dotted line).}
\end{figure}

Fig. 7 shows a model that uses a 3\textsuperscript{rd} degree
polynomial. As one can see, the model follows the empirical data
sufficiently yet not optimally, which is confirmed by a relatively
decent fit of the model, \emph{R}\textsuperscript{2} = 0.885 at p
\textless{} 0.05 (and 0.655 for the unweighted model). Certainly,
higher-order polynomials would lead to a better fit (and ultimately to
overfitting), but at the same time all these complex models have to be
rejected because of their high \emph{p}-values. Still, even the
3\textsuperscript{rd} degree polynomial does not seem to explain the
dispersion of data points in the years 1450--1550. Note, however, that
after 1550 the course of the change follows the classical logistic
curve: if we only took into account the data from this year and used the
classical model, the value of \emph{R}\textsuperscript{2} would increase
substantially.

\hypertarget{wszytek-wszystek}{%
\subsection{\texorpdfstring{\emph{wszytek} \textgreater{}
\emph{wszystek}}{wszytek \textgreater{} wszystek}}\label{wszytek-wszystek}}

The change \emph{wszytek \textgreater{} wszystek} `all' has already been
discussed in literature (Michalska, 2013; Klemensiewicz, 1965), yet with
no quantitative evidence taken into account. Interestingly, large-scale
material excerpted from our diachronic corpus does not exhibit a clear
picture either. To be precise, it is difficult to identify a clean
Piotrowski's process here (Fig. 8). Although the trend stabilizes in the
second half of the 18\textsuperscript{th} century and is substantially
closer to the expected ``s''-shaped curve, for the most part the results
are very blurred. Even the use of a polynomial logistic regression model
does not help much: despite the decent degree of model fit,
\emph{R}\textsuperscript{2} = 0.784 when using a 3\textsuperscript{rd}
degree polynomial, it can be easily seen that for the period covering
the 15\textsuperscript{th}--17\textsuperscript{th} centuries, the
distribution of points does not resemble the modeled curve. Fitting to
empirical data can be artificially increased by using a model with a
higher degree of the polynomial (e.g.~for a 4\textsuperscript{th} degree
polynomial \emph{R}\textsuperscript{2} = 0.837), nonetheless, such a
model doesn't show statistical significance any more (the \emph{p}-value
increases well beyond the 0.05 level), which simply means that the model
does not sufficiently describe the data.

\begin{figure}
\centering
\includegraphics[width=1\textwidth]{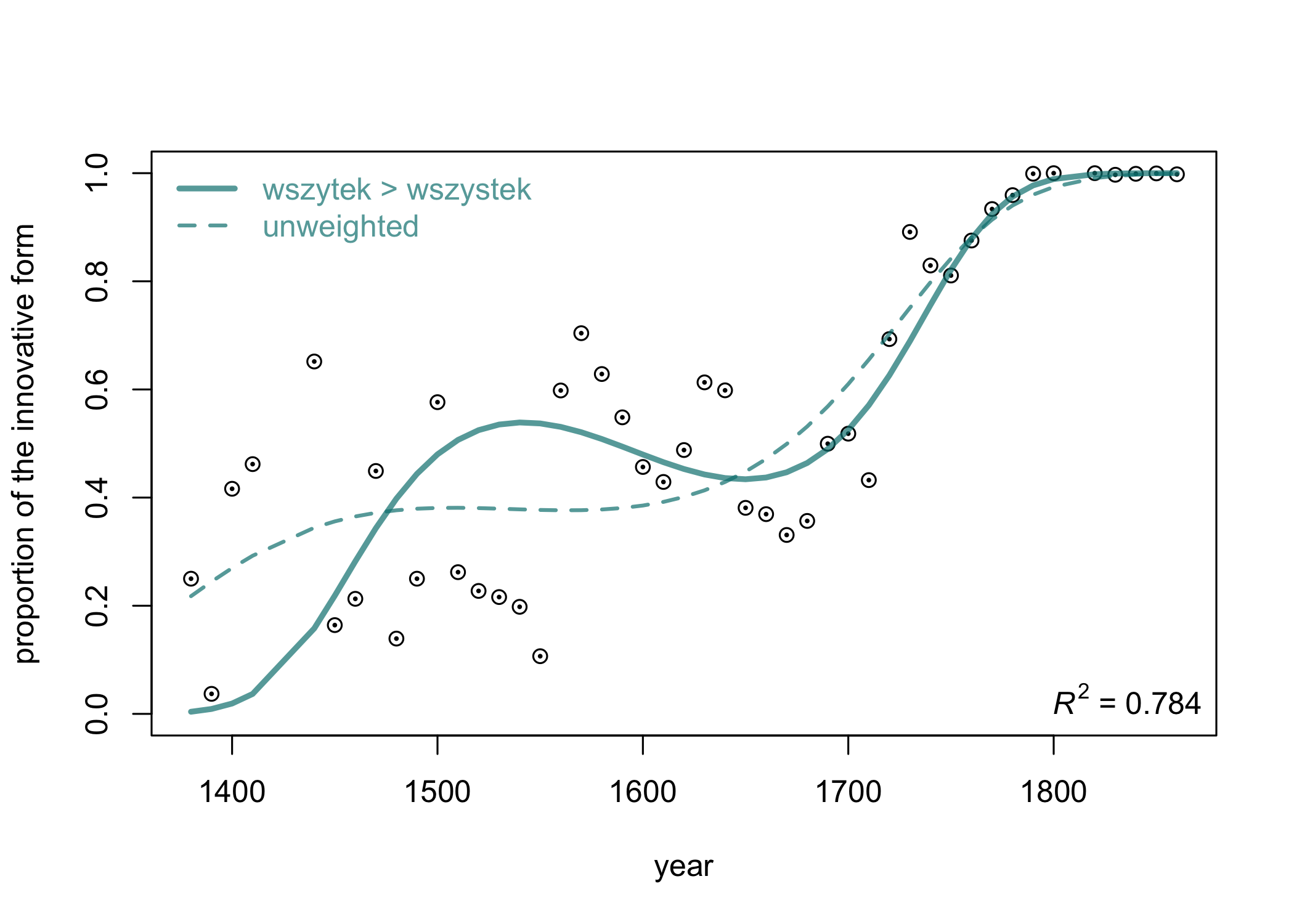}
\caption{The course of change \emph{wszytek} \textgreater{}
\emph{wszystek} (in all inflecting forms).}
\end{figure}

Again, an immediate observation can be made about the validity of
Piotrowski's ``law''. This time, however, data scarcity cannot be blamed
for the remarkable dispersion of the data points spanning over the
14\textsuperscript{th}--17\textsuperscript{th} centuries. The change
\emph{wszytek \textgreater{} wszystek} with its 55,005 attestations is
our best documented case.

\hypertarget{competing-forms-abo-and-albo}{%
\subsection{\texorpdfstring{Competing forms \emph{abo} and
\emph{albo}}{Competing forms abo and albo}}\label{competing-forms-abo-and-albo}}

In the case of \emph{abo} and \emph{albo} `or' we don't observe a
transition from one form to another, but rather a competition of two
forms. In the Middle Ages, \emph{albo} was typical for southern Poland,
whereas the northern regions adhered to \emph{abo} (Michalska, 2013).
Moreover, the latter form was considered to belong to lower registers --
hence its absence in religious texts. In the 16\textsuperscript{th}
century \emph{abo} percolates to all regions and all genres. It has also
been suggested that the spread of \emph{abo} in high brow texts is due
to a typo mistake in the 1560 edition of the very popular \emph{Postil}
by Mikołaj Rej (Urbańczyk, 1953). Whatever the actual reason of the
change might be, important in the context of our study is the
accelerating growth of \emph{abo} at the cost of \emph{albo}. In the
first centuries the trajectory resembles Piotrowski's process. Just a
few decades before winning the game, however, the form \emph{abo}
unexpectedly derails: starting from the mid-17\textsuperscript{th}
century, the almost-extinct form \emph{albo} gradually regains its
position of a ruler, eventually pushing \emph{abo} into the status of
archaism in the 19\textsuperscript{th} century.

Using a standard logistic model would not yield any acceptable results
in this case, since the curve (see Fig. 9) does hardly resemble the
shape of an ``s''. Instead, the polynomial logistic regression model
seems to better describe such a non-obvious competition of two forms.
The goodness-of-fit is relatively high, \emph{R}\textsuperscript{2} =
0.868 when a 6\textsuperscript{th} degree polynomial is used
(\emph{R}\textsuperscript{2} = 0.803 for the unweighted model). Although
the model is rather complex, with multiple parameters, and difficult to
defend against Occam's razor, it nevertheless explains the empirical
data decently.

\begin{figure}
\centering
\includegraphics[width=1\textwidth]{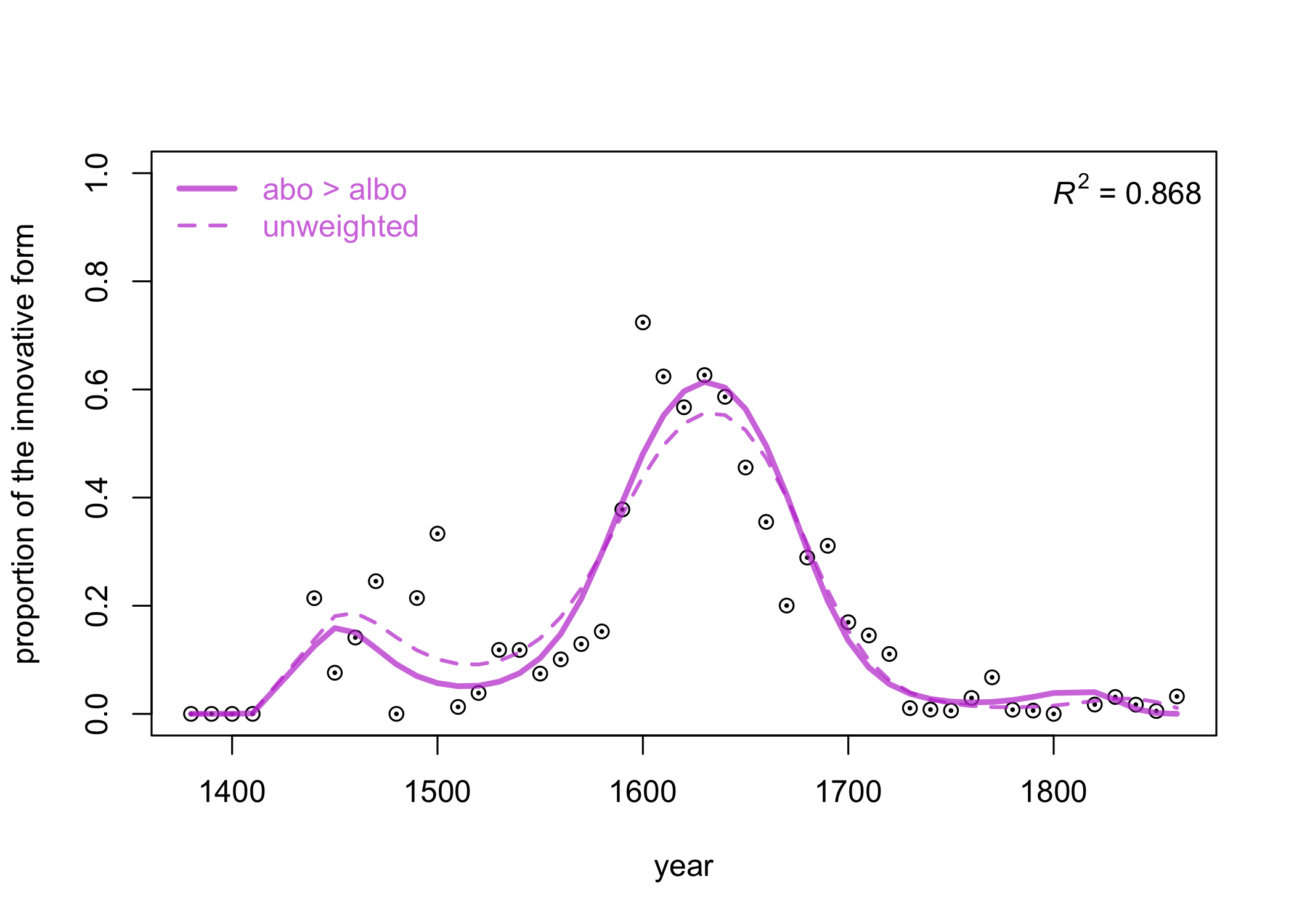}
\caption{The competing forms \emph{abo} and \emph{albo}: a model
of polynomial logistic regression.}
\end{figure}

From a theoretical point of view, the competition of \emph{abo} and
\emph{albo} can also be viewed as a combination of two logistic curves,
one of which would mirror the other: ideally, a raising curve (which can
be imagined as the lower part of the letter ``s'') would smoothly evolve
into its mirror image, or a descending curve, as was proposed by
Vulanović (2007: 113). Fig. 10 shows exactly the same dataset for
\emph{abo} and \emph{albo}, but this time it has been split into two
parts and modeled independently of each other, using classical logistic
regression. As it turns out, the first part (1380--1610) fits to the
model to a much lesser extent (\emph{R}\textsuperscript{2} = 0.47) than
the second part (\emph{R}\textsuperscript{2} = 0.921). Also, the
hypothesis of two mirror logistic curves finds no confirmation in our
data: in different scenarios that we tested, the two best-fit models
never matched, no matter where the split into the two parts was applied.

\begin{figure}
\centering
\includegraphics[width=1\textwidth]{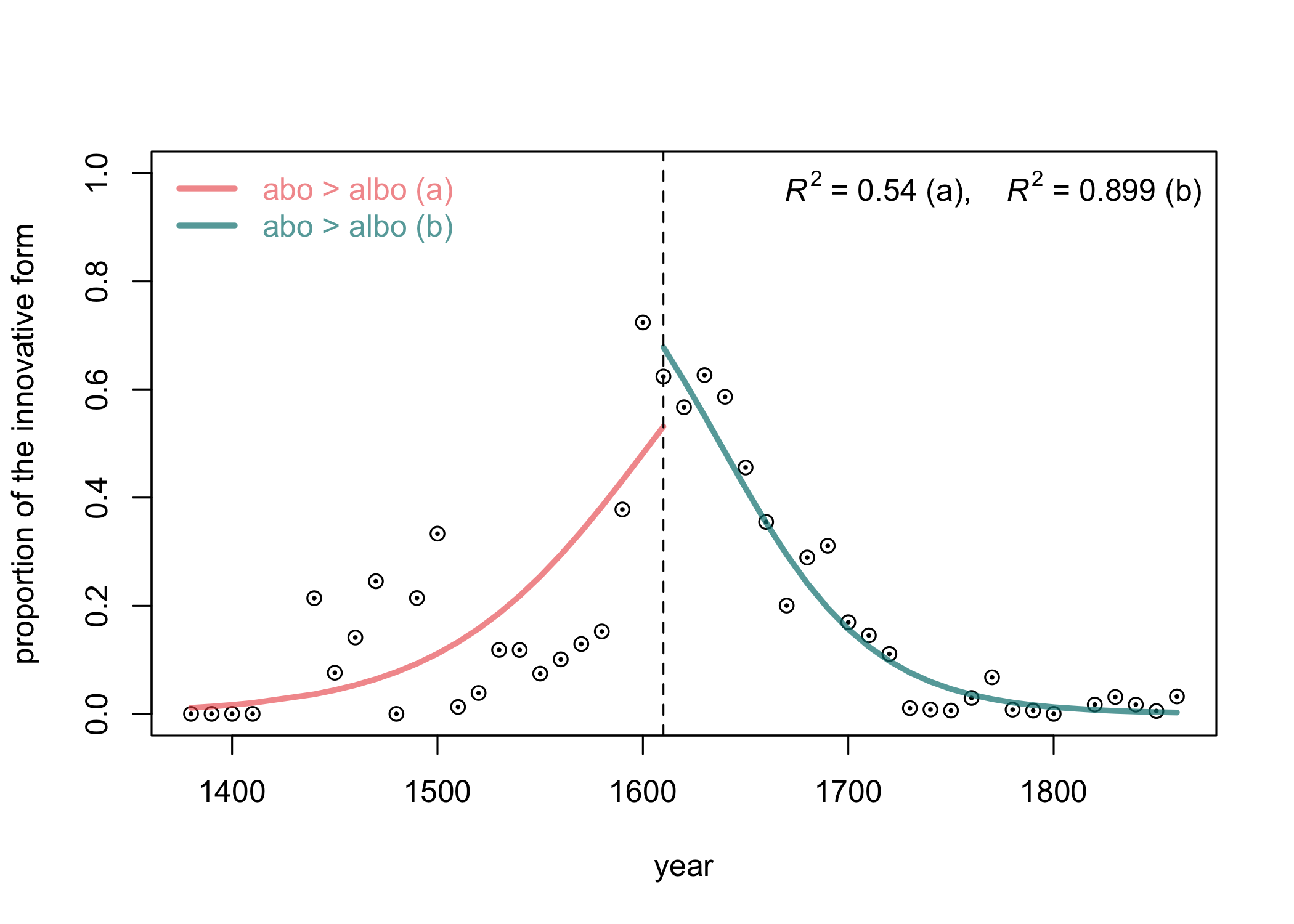}
\caption{The competing forms \emph{abo} and \emph{albo}: two
independent logistic models computed separately for two time spans,
1380--1610 and 1610--1850.}
\end{figure}

The results shown in Figs. 9 and 10 might give an impression that they
describe two substantially different courses of change \emph{abo}
\textgreater{} \emph{albo}. However, this is only an illusion, as the
data points (black dots) in both plots are identical. Different varieties
of lines, colors or shapes superimposed on a graph strongly affect our
perception of actual data -- the effect is known as visual rhetoric
bias. This leads us to a two-fold observation. Firstly, the models
discussed above reconstruct different trajectories of change -- in our
situation it is either a two-hump run (Fig. 9) or a one-hump run (Fig.
10) -- and it is not obvious at all which of the two models is more
likely to tell the truth. Secondly, the eyeballing is prone to bias --
as we have seen above -- hence the need to supplement the plots with the
goodness-of-fit measures, such as \emph{R}\textsuperscript{2} or one of
its variants.

\hypertarget{the-dynamics-of-language-change}{%
\subsection{The dynamics of language
change}\label{the-dynamics-of-language-change}}

Given the above results discussed individually, a natural question
emerges of how the particular changes relate one to another. Do they
travel through time in a collinear fashion? Although we have no reason
to suspect a perfect overlap of the changes nor a particular order (the
changes are not caused one by another), yet still, the question remains
as to the general dynamics of language change. Do all the changes
advance at the same pace? It has already been suggested that a few
individual logistic curves might be combined into a single
multidimensional model (Vulanović, 2012). However, in our study we
propose a somewhat simpler approach, namely we suggest to focus first on
a composite picture of all the trajectories merged into one plot (Fig.
11), which shows a gradual accumulation of several individual
(independent) changes.

\begin{figure}
\centering
\includegraphics[width=1\textwidth]{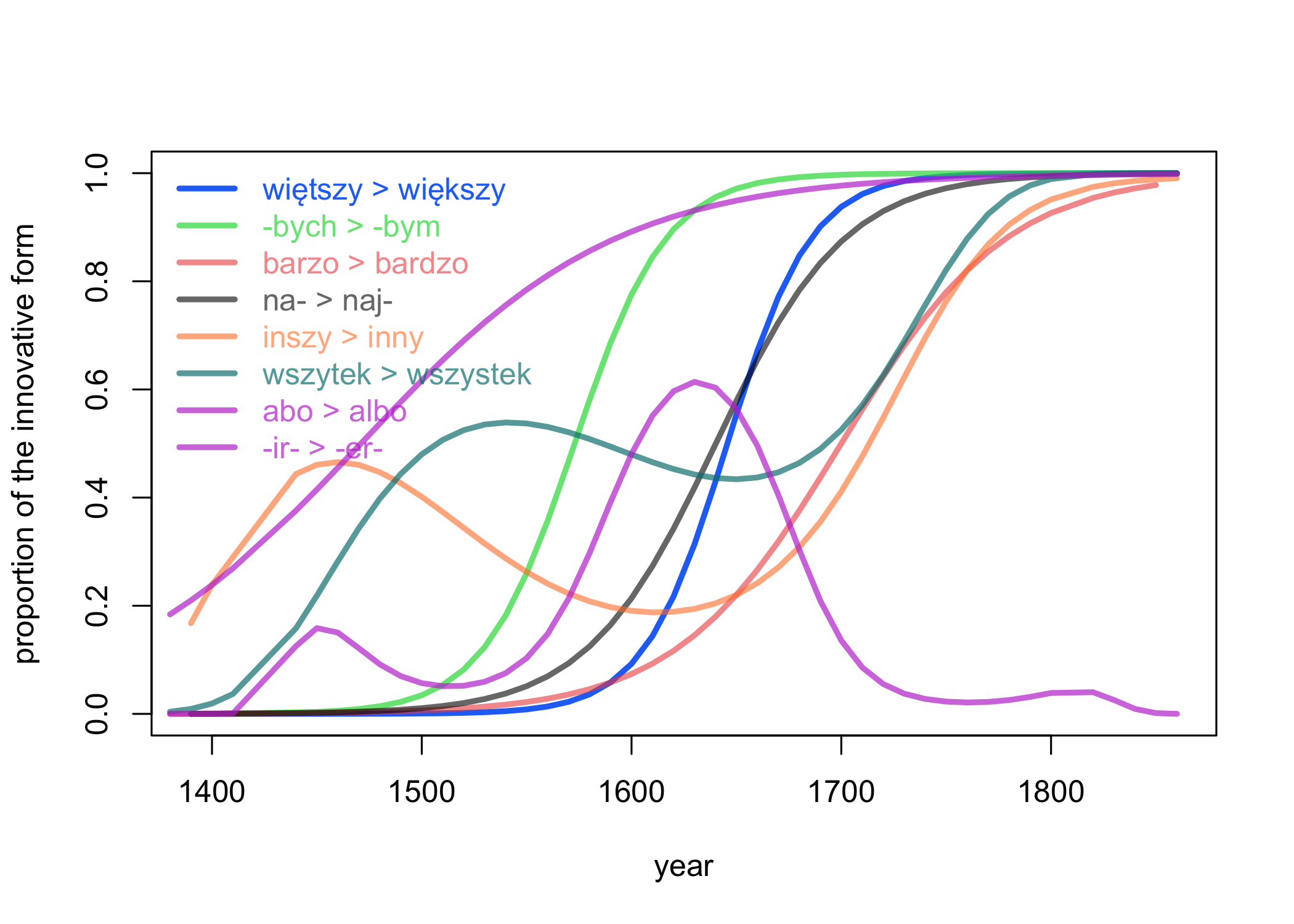}
\caption{The overall picture of the language changes in Middle
Polish.}
\end{figure}

The complex symphony of individual models represented in one plot shows
how different the dynamics of the diachronic process can be. While
\emph{na-} \textgreater{} \emph{naj-} is a slow, time-stretched process,
then \emph{-bych} \textgreater{} \emph{-bym} is a sudden change, so is
also \emph{więtszy} \textgreater{} \emph{większy}. What makes these
processes different, is the fact that they happen at a different time --
when the probability of the innovative \emph{-bym} is already close to
1, the probability to find the innovative form \emph{większy} is barely
reaching 0.5. To put it simply: when the former process is about halfway
through, the latter is at its beginning. If, on the other hand, we
compare \emph{-bych} \textgreater{} \emph{-bym} to \emph{barzo}
\textgreater{} \emph{bardzo}, we find that the second process is even
more recent (it starts when \emph{-bych} already disappears from the
system) and meanwhile somewhat slower, which can be seen from the more
gentle slope of the modeled line.

Since one could expect the changes to occur at different times, the lack
of any collinearities or consistent breakthroughs in, say,
16\textsuperscript{th} or 17\textsuperscript{th} century, is rather
surprising. If such a collinearity had surfaced, if would have supported
the hypothesis that a language change is a combination of several
variables occurring roughly at the same time. Our results suggest
otherwise that it is difficult to identify a clear-cut moment of a
general language change; rather, we deal here with a gradual
accumulation of several individual (independent) changes.

\hypertarget{evaluation}{%
\section{Evaluation}\label{evaluation}}

No matter how convincing the presented results might seem, a crucial
question has to be asked: to which extent are these results stable and
reproducible? In the following section, we will address the issue by a
systematic examination of the conditions of our experimental setup.

There are at least three elements of the procedure that might have
affected the final model. Firstly, since the shape of the logistic curve
is a derivative of the estimated parameters \(\beta_0\) and \(\beta_1\),
one could ask whether any distortions might appear at this stage of
modeling. Arguably, there is no room for manipulation here: the
parameters \(\beta_0\) and \(\beta_1\) are estimated by solving an
equation that minimizes the difference between the model and empirical
data (James et al., 2013: 133--34). There exists only one pair of values
that satisfies the equation, therefore it is not possible for the result
to be tweaked at this stage. A given set of input values will always
give one optimal solution, even if the curve does not look optimal to
the human eye.

The second source of distortions is, obviously, the input dataset
itself: it is rather naïve to expect that solid results could be
obtained from a corpus of poor quality. Although this component has an
immense impact on the entire experiment, there was little we could do
about it, apart from collecting as many texts as possible and checking
the individual texts for quality. Still, we assume that the amount of
systematic error we potentially deal with, is an inherent feature of the
dataset and does not change from one experiment to another.

Finally, the third source of possible distortions is our procedure of
dividing the data into smaller subcorpora, as discussed above in the
Method section. Analyzing diachronic phenomena always requires the
dataset to be divided into smaller time spans, therefore no objections
can be raised as to the idea of splitting. What calls for explanation,
however, is the size of a single time span. As mentioned above, the
chosen subcorpus size should account for data scarcity, while keeping
the noise at the lowest possible level.

While the choice of the subcorpus size might be -- and usually is --
somewhat arbitrary, the size of the overlap between particular
subcorpora is even more so. Again, we can assume that a large overlap
size (i.e.~a slow pace of the moving window) results in smoother data
points, but it comes at the cost of their reliability, especially when a
single outlier falls into several subcorpora at a time. Due to the lack
of strong theoretical assumptions, both the subcorpus size and the
overlap have to be decided by choosing the combination of parameters
that yield best results.

We approached the above issue empirically, by testing numerous possible
combinations of the subcorpus size and the overlap in a grid-search
manner. We conducted a series of tests for 20 different sizes of the
subcorpora: 5, 10, 15, 20, \ldots, 100 years, and for 20 different sizes
of the overlap, modeling the 7 linguistic changes discussed above (we
excluded \emph{abo} \textgreater{} \emph{albo}, and merged \emph{-bych}
\textgreater{} \emph{-bym} and \emph{-bychmy} \textgreater{}
\emph{-byśmy} into one instance) for a total of 2,400 models for
comparison. In fact, we were able to compute the parameters for only
1,200 models, since the overlap size cannot exceed the current subcorpus
size. For each of the models, we calculated the
\emph{R}\textsuperscript{2} coefficient of goodness-of-fit. {We repeated
the entire procedure for both weighted and classical logistic regression
models. Due to the fact that the general picture turned out to be
similar in these two groups of models, below we report the
\emph{R}\textsuperscript{2} values for the unweighted variant only.} A
small portion of results for the 50-year subcorpora with an overlap of
20 years, and the 20-year subcorpus with an overlap of 5 years, are
presented in Table 1.

\begin{longtable}[]{@{}lrrr@{}}
\toprule
change & subcorpus 50 years & subcorpus 20 years &\tabularnewline
\midrule
\endhead
\emph{więtszy} \textgreater{} \emph{większy} & 0.952 & 0.921
&\tabularnewline
-\emph{bych} \textgreater{} -\emph{bym}\footnote{Including the plural
  forms -\emph{bychmy} \textgreater{} -\emph{byśmy}.} & 0.983 & 0.954
&\tabularnewline
\emph{barzo} \textgreater{} \emph{bardzo} & 0.903 & 0.818
&\tabularnewline
\emph{na}- \textgreater{} \emph{naj}- & 0.950 & 0.898 &\tabularnewline
\emph{inszy} \textgreater{} \emph{inny} & 0.937 & 0.887 &\tabularnewline
\emph{wszytek} \textgreater{} \emph{wszystek} & 0.856 & 0.779
&\tabularnewline
-\emph{ir}- \textgreater{} -\emph{er}- & 0.647 & 0.608 &\tabularnewline
\bottomrule
\end{longtable}

Table 1. Goodness-of-fit of the logistic model to empirical data
(McFadden's pseudo-\emph{R}\textsuperscript{2}) depending on different
ways of dividing data into subcorpora.

With the large number of results for 1,200 independent test, discussing
them individually would be difficult, therefore we present them in a
compact form in Figs. 12--13. The plots show the goodness-of-fit as a
function of the subcorpus size, for a 10-year overlap and for a 20-year
overlap, respectively. Relatively high \emph{R}\textsuperscript{2} for
big subcorpora (say, of 50 or 100 years) comes at no surprise given the
smaller number of degrees of freedom and the smoothing effect of large
moving windows, but the actual goal here is to decrease the subcorpus
size as far as possible while still observing acceptable results. The
general picture is fairly optimistic, since we don't encounter any
substantial decrease of \emph{R}\textsuperscript{2} for subcorpora
between 20 and 100 years. Below the value of 20 years, the performance
of modeling decreases significantly.

\begin{figure}
\centering
\includegraphics[width=1\textwidth]{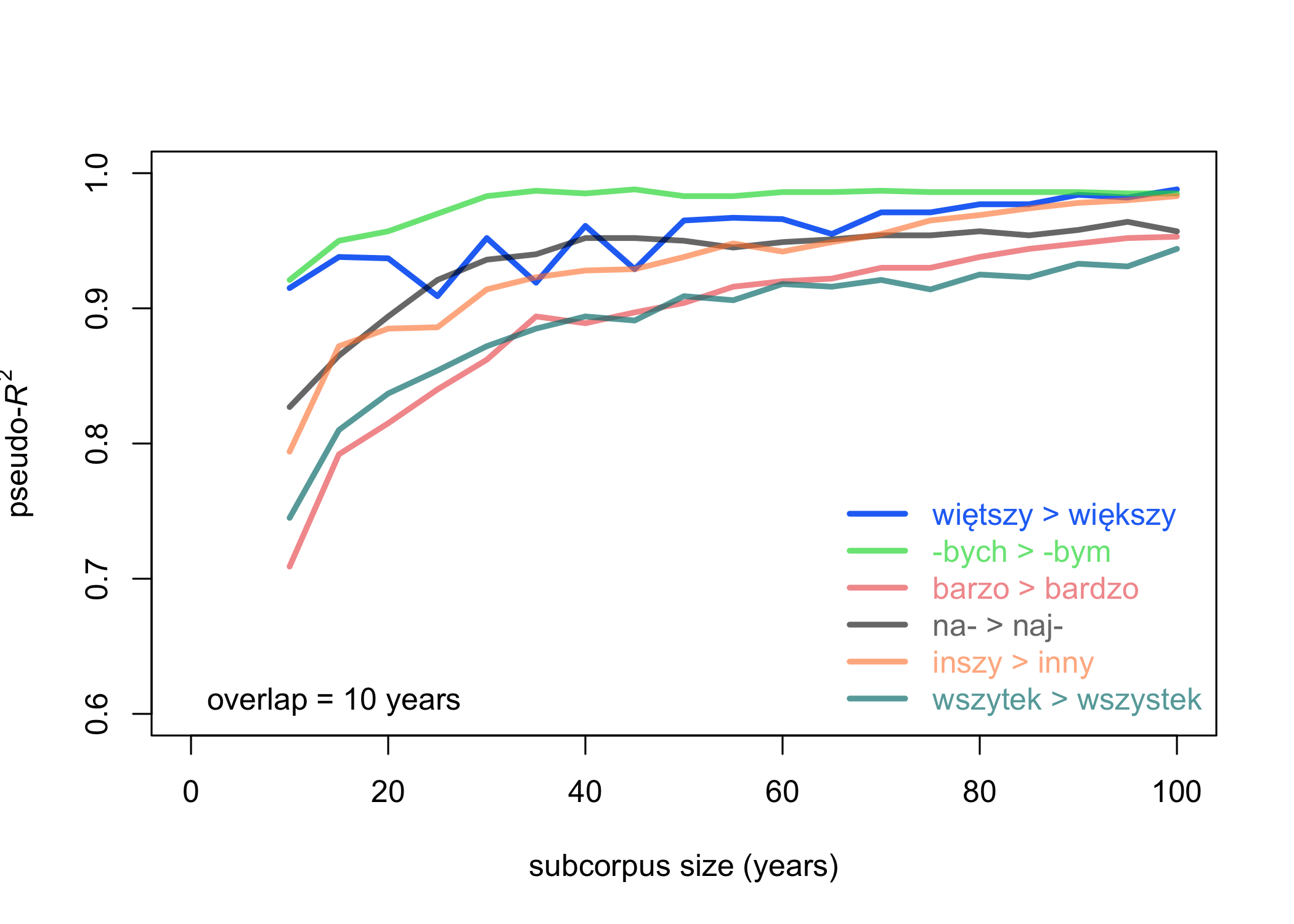}
\caption{The influence of the subcorpus size on the goodness of
fit (10-year overlap).}
\end{figure}

When it comes to the overlap size, the differences between Fig. 12 and
Fig. 13 clearly show that a small overlap -- which can be also
interpreted as a fast pace of the moving window -- produces smoother
results than slowly moving windows. Arguably, we are observing here an
effect of the outliers falling into several subcorpora when the window
progresses too slowly.

\begin{figure}
\centering
\includegraphics[width=1\textwidth]{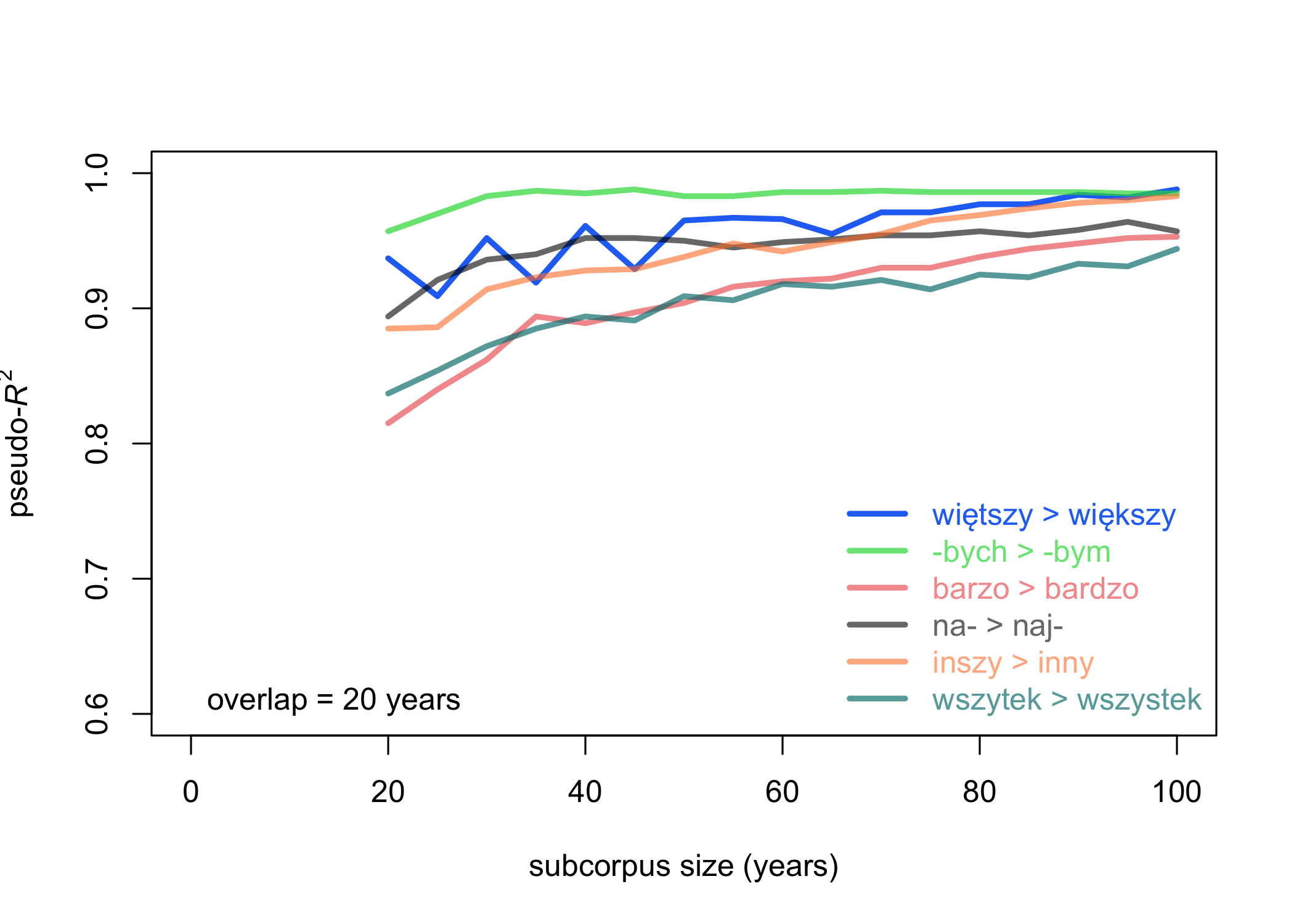}
\caption{The influence of the subcorpus size on the goodness of
fit (20-year overlap).}
\end{figure}

The evaluation tests allow us to draw a few conclusions. Firstly, the
relatively stable \emph{R}\textsuperscript{2} across the board might
suggest that the underlying logistic processes for our discussed changes
are perhaps stronger than we suspect; apparently, dividing the corpus
arbitrarily into any subcorpora of 20 or more years would still yield
satisfactory results. Secondly, thanks to the above systematic tests we
have gain empirical justification for the arbitrary size of the
subcorpus discussed in the previous sections, namely 20 years with a
10-year overlap.

\hypertarget{discussion}{%
\section{Discussion}\label{discussion}}

The results presented in this study exhibit a reasonable predictive
power, namely the goodness-of-fit for particular logistic regression
models turned out to be rather high. The best fit reached
\emph{R}\textsuperscript{2} = 0.957, and rarely dropped below 0.7. There
is a fly in the ointment, though. No matter how many cases follow the
expected course of change, the exceptions from the model cannot be
denied. Sooner or later, one arrives at a case when the observed results
do not resemble any ``s''-shaped trajectory and yet, insufficient amount
of data cannot be blamed for the effect. It is true that sometimes a
more complex variants of the logistic models can yield reasonable
results. However, this hardly satisfies a historical linguist, who might
want to ask: Why do some changes follow the model, while others don't?
Or perhaps a valid question is the opposite one: why should we expect
the changes to always follow the logistic curve? Language can be
affected by several -- internal and external -- factors; a conservative
school education system can slow down the changes, and the same goes for
the prescriptive linguistics' authority, whereas progressive literary
influencers or, in more recent times, social media can accelerate the
changes substantially. Why wouldn't other variables play role in this
fairly complex picture?

The reservations formulated above shed some light on the nature of
language changes examined in this study. We refer to them as the
\emph{so-called} Piotrowski's law, because we hardly believe one deals
here with any linguistic quantitative law in the sense of Zipf's,
Menzerath's or Heaps's law. For in the classical sense, a quantitative
linguistic law in order to be named as such must occur under all
conditions and in every language -- it must be universal. In contrast,
the process observed by Piotrowski applies only to a certain group of
linguistic changes. Therefore, we rather agree with Stachowski that such
a non-linear process does not actually apply to language and its
self-organization (as many other laws of quantitative linguistics do),
but to social processes related to language (Stachowski, 2020).

\hypertarget{conclusion}{%
\section{Conclusion}\label{conclusion}}

In this paper, we examined a few changes in Polish over the relatively
wide time span of 15\textsuperscript{th}--19\textsuperscript{th}
centuries, some of them being successful realizations of the so-called
Piotrowski's law. In order to quantify these cases, we used logistic
regression modeling, followed by the \emph{p}-value and the
\emph{R}\textsuperscript{2} coefficient to assess the model's
goodness-of-fit. We also extended the standard approach by using general
logistic regression models with polynomial kernels, for the cases which
hardly resemble Piotrowski's process. Yet, even the polynomial models
could not explain the changes that exhibited a considerable dispersion
of data points. This oftentimes applied to early periods of Polish,
whereas the evidence for 17\textsuperscript{th}--19\textsuperscript{th}
centuries usually showed a better match with the models.

The set of changes which we have chosen for our study are documented in
a single corpus, and they run concurrently. This offers a unique
opportunity not only for comparing the fit of the models to each dataset
individually, but also to observe how they relate one to another. Our
results clearly show that the changes that have been traditionally (yet
tacitly) assumed to exhibit similar dynamics, are in fact very different
in their trajectories.

The observation that needs to be commented on, is that some of the
changes are not compliant with the model or, to be precise, with the
standard logistic regression model of phase change. Since the change
\emph{abo} \textgreater{} \emph{albo} can be explained by a failed
innovation scenario, the other irregular changes don't fall into this
category at all. Moreover, in some cases the blurry picture cannot be
simply explained by data scarcity. Presumably, our results show once
more that the logistic ``s''-shaped trajectory, though it describes most
of the processes to an acceptable degree, cannot be considered a
universal underlying rule for the language change. While some exceptions
from the so-called Piotrowski's law have been already reported in the
literature, our study provides new and rich evidence of this phenomenon.

Last but not least, our experiment shows that even a relatively small
diachronic corpus (12 million words in total) is large enough to trace
language changes over the course of a few centuries. This is a fairly
optimistic outcome, because diachronic corpora will always suffer from
insufficient coverage and the lack of balance. We would like to hope
that our findings can serve as an encouraging example for further
investigations.

\hypertarget{acknowledgements}{%
\section{Acknowledgements}\label{acknowledgements}}

This research was conducted as a result of a project supported by
Poland's National Science Centre (project number
UMO-2013/11/B/HS2/02795).

\hypertarget{references}{%
\section*{References}\label{references}}
\addcontentsline{toc}{section}{References}

\hypertarget{refs}{}
\begin{CSLReferences}{1}{0}
\leavevmode\hypertarget{ref-altmann_piotrowski-gesetz_1983-1}{}%
\textbf{Altmann, G.} (1983). Das {Piotrowski}-{Gesetz} und seine
{Verallgemeinerungen}. In Best, K.-H. and Kohlhase, J. (eds),
\emph{Exakte {Sprachwandelforschung}. {Theoretische} {Beiträge},
Statistische {Analysen} Und {Arbeitsberichte}}. Göttingen: Edition
Herodot, pp. 54--90.

\leavevmode\hypertarget{ref-basu_ill_2016}{}%
\textbf{Basu, A.} (2016). {`{Ill} shapen sounds, and false
orthography'}: {A} computational approach to {Early} {English}
orthographic variation. In Ullyot, M., Jakacki, D. and Estill, L. (eds),
\emph{Early Modern Studies and the Digital Turn}. (New {Technologies} in
{Medieval} and {Renaissance} {Studies} 6). Toronto \& Tempe: Iter \&
ACMRS, pp. 169--201.

\leavevmode\hypertarget{ref-best_zum_1983-1}{}%
\textbf{Best, K.-H.} (1983). Zum morphologischen {Wandel} einiger
deutscher {Verben}. In Best, K.-H. and Kohlhase, J. (eds), \emph{Exakte
{Sprachwandelforschung}. {Theoretische} {Beiträge}, Statistische
{Analysen} Und {Arbeitsberichte}}. Göttingen: Edition Herodot, pp.
107--18.

\leavevmode\hypertarget{ref-best_iranismen_2013}{}%
\textbf{Best, K.-H.} (2013). Iranismen im deutschen.
\emph{Glottometrics}, \textbf{26}: 1--8.

\leavevmode\hypertarget{ref-gnatchuk_anglicisms_2015}{}%
\textbf{Gnatchuk, H.} (2015). Anglicisms in the {Austrian} newspaper
{`{Kleine} {Zeitung}'}. \emph{Glottometrics}, \textbf{31}: 38--49.

\leavevmode\hypertarget{ref-gorski_zmiana_2019}{}%
\textbf{Górski, R. L., Król, M. and Eder, M.} (2019). \emph{Zmiana w
języku. Studia kwantytatywno-korpusowe}. {Kraków}: {IJP PAN}.

\leavevmode\hypertarget{ref-james_introduction_2013}{}%
\textbf{James, G., Witten, D., Hastie, T. and Tibshirani, R.} (2013).
\emph{An Introduction to Statistical Learning with Applications in {R}}.
New York: Springer.

\leavevmode\hypertarget{ref-klemensiewicz_historia_1965}{}%
\textbf{Klemensiewicz, Z.} (1965). \emph{Historia języka polskiego}.
Warszawa: PWN.

\leavevmode\hypertarget{ref-kowalska_rozwoj_1978}{}%
\textbf{Kowalska, A.} (1978). Rozwój nowych form słowa posiłkowego:
jestem, jesteś, jesteśmy, jesteście. \emph{Poradnik Językowy},
\textbf{9}: 377--84.

\leavevmode\hypertarget{ref-kohler_tuzzi_linguistic_2015}{}%
\textbf{Köhler, R. and Tuzzi, A.} (2015). Linguistic modelling of
sequential phenomena. In Mikros, G. K. and Macutek, J. (eds),
\emph{Sequences in Language and Text}. Walter de Gruyter, pp. 109--23.

\leavevmode\hypertarget{ref-leopold_piotrowski-gesetz_2005-1}{}%
\textbf{Leopold, E.} (2005). Das {Piotrowski}-{Gesetz}. In Köhler, R.,
Altmann, G. and Piotrowski, R. (eds), \emph{Quantitative {Linguistik} --
{Quantitative} {Linguistics}. {Ein} Internationales {Handbuch}}. Berlin
/ New York: de Gruyter, pp. 627--33.

\leavevmode\hypertarget{ref-mcfadden_conditional_1973}{}%
\textbf{McFadden, D.} (1973). Conditional logit analysis of qualitative
choice behavior. In Zarembka, P. (ed), \emph{Frontiers in Econometrics}.
{New York}: {Academic Press}, pp. 105--42.

\leavevmode\hypertarget{ref-meyer_english_2002}{}%
\textbf{Meyer, C. F.} (2002). \emph{English Corpus Linguistics: {An}
Introduction}. {Cambridge}: {Cambridge University Press}.

\leavevmode\hypertarget{ref-michalska_status_2013-1}{}%
\textbf{Michalska, P.} (2013). \emph{Status staropolskich oboczności
wyrazowych w polszczyźnie doby średniopolskiej}. Poznań: Wydawnictwo
PTPN.

\leavevmode\hypertarget{ref-motyl_normalizacja_2014}{}%
\textbf{Motyl, A.} (2014). \emph{Normalizacja fleksji werbalnej w
zakresie kategorii czasu w dobie średniopolskiej}. Poznań: Wydawnictwo
PTPN.

\leavevmode\hypertarget{ref-piotrowskaja_matematiceskie_1974}{}%
\textbf{Piotrowskaja, A. and Piotrowski, R.} (1974). Matematičeskie
modeli v diachronii i tekstoobrazovanii. In Piotrowski, R. (ed),
\emph{Statistika reči i avtomatičeskij analiz teksta}. Leningrad: Nauka,
pp. 361--400.

\leavevmode\hypertarget{ref-stachowski_german_2016}{}%
\textbf{Stachowski, K.} (2016). German loanwords in {Polish} and remarks
on the {Piotrowski}-{Altmann} law. \emph{Issues in {Quantitative}
{Linguistics} 4}. Lüdenscheid: RAM-Verlag, pp. 237--59.

\leavevmode\hypertarget{ref-stachowski_piotrowski-altmann_2020}{}%
\textbf{Stachowski, K.} (2020). Piotrowski-{Altmann} law: {State} of the
art. \emph{Glottotheory}, \textbf{11}(1): 3--14.

\leavevmode\hypertarget{ref-taszycki_staropolskie_1946}{}%
\textbf{Taszycki, W.} (1946). Staropolskie formy czasu przeszłego
{`robiłech'}, {`robilichmy'}. \emph{Sprawozdania {Polskiej} {Akademii}
{Umiejętności}}, vol. 46. pp. 7--10.

\leavevmode\hypertarget{ref-urbanczyk_rola_1953}{}%
\textbf{Urbańczyk, S.} (1953). \emph{Rola wielkich pisarzy odrodzenia na
tle innych czynników kształtujących język literacki}. Warszawa: Polska
Akademia Nauk.

\leavevmode\hypertarget{ref-vulanovic_fitting_2007-1}{}%
\textbf{Vulanović, R.} (2007). Fitting periphrastic {``do''} in
affirmative declaratives. \emph{Journal of Quantitative Linguistics},
\textbf{14}(2-3): 111--26.

\leavevmode\hypertarget{ref-vulanovic_multidimensional_2012}{}%
\textbf{Vulanović, R.} (2012). A multidimensional generalization of the
{Piotrowski}-{Altmann} law. \emph{Methods and {Applications} of
{Quantitative} {Linguistics}: {Selected} Papers of the 8th
{International} {Conference} on {Quantitative} {Linguistics}
({QUALICO})}. Belgrade, pp. 184--93.

\leavevmode\hypertarget{ref-vulanovic_fitting_2007}{}%
\textbf{Vulanović, R. and Baayen, H.} (2007). Fitting the development of
periphrastic {``do''} in all sentence types. In Grzybek, P. and Köhler,
R. (eds), \emph{Exact Methods in the Study of Language and Text:
Dedicated to {Gabriel} {Altmann} on the Occasion of His 75th Birthday}.
Berlin: de Gruyter, pp. 679--88.

\end{CSLReferences}

\bibliographystyle{unsrt}
\bibliography{bibliography.bib}

\end{document}